\documentclass[journal,comsoc]{IEEEtran}
\usepackage{amsmath,amsfonts}
\usepackage{algorithmic}
\usepackage[linesnumbered,ruled,vlined]{algorithm2e}
\usepackage{array}
\usepackage[caption=false,font=normalsize,labelfont=sf,textfont=sf]{subfig}
\usepackage[justification=centering]{caption}
\usepackage{epsfig,amssymb,bm,dsfont}
\usepackage{textcomp}
\usepackage{stfloats}
\usepackage{multirow}
\usepackage{url}
\usepackage{verbatim}
\usepackage{graphicx}
\usepackage{cite}
\usepackage{bbm} 
\allowdisplaybreaks

\newtheorem{theorem}{\textbf{Theorem}}

\newtheorem{lemma}{\textbf{Lemma}}

\begin{document}

\title{Trustworthy DNN Partition for Blockchain-enabled Digital Twin in Wireless IIoT Networks}

\author{Xiumei~Deng,~Jun~Li,~Long~Shi,~Kang~Wei,~Ming~Ding,~Yumeng~Shao,~Wen~Chen,~and~Shi~Jin
	\thanks{X. Deng, J. Li, Y. Shao and L. Shi are with School of Electronic and Optical Engineering, Nanjing University of Science and Technology, Nanjing 210094, China. E-mail: \{xiumeideng, jun.li, yumengshao\}@njust.edu.cn, slong1007@gmail.com.}
	\thanks{K. Wei is with the Department of Computing, Hong Kong Polytechnic University, Hong Kong 999077, China. E-mail: kangwei@polyu.edu.hk.}
	\thanks{M. Ding is with Data61, CSIRO, Sydney, NSW 2015, Australia. E-mail: ming.ding@data61.csiro.au.}
	\thanks{W. Chen is with Department of Electronics Engineering, Shanghai Jiao Tong University, Shanghai 200240, China. E-mail: wenchen@sjtu.edu.cn.}
	\thanks{S. Jin is with the National Mobile Communications Research Laboratory, Southeast University, Nanjing 210096, China. E-mail: jinshi@seu.edu.cn.}}

\markboth{}%
{Shell \MakeLowercase{\textit{et al.}}: A Sample Article Using IEEEtran.cls for IEEE Journals}


\maketitle

\begin{abstract}
Digital twin (DT) has emerged as a promising solution to enhance manufacturing efficiency in industrial Internet of Things (IIoT) networks. To promote the efficiency and trustworthiness of DT for wireless IIoT networks, we propose a blockchain-enabled DT (B-DT) framework that employs deep neural network (DNN) partitioning technique and reputation-based consensus mechanism, wherein the DTs maintained at the gateway side execute DNN inference tasks using the data collected from their associated IIoT devices. First, we employ DNN partitioning technique to offload the top-layer DNN inference tasks to the access point (AP) side, which alleviates the computation burden at the gateway side and thereby improves the efficiency of DNN inference. Second, we propose a reputation-based consensus mechanism that integrates Proof of Work (PoW) and Proof of Stake (PoS). Specifically, the proposed consensus mechanism evaluates the off-chain reputation of each AP according to its computation resource contributions to the DNN inference tasks, and utilizes the off-chain reputation as a stake to adjust the block generation difficulty. In this way, the degradation of PoW consensus security is compensated by off-chain reputation stake, which improves the consensus efficiency while ensuring the trustworthiness on the chain. Third, we formulate a stochastic optimization problem of communication resource (i.e., partition point) and computation resource allocation (i.e., computation frequency of APs for top-layer DNN inference and block generation) to minimize system latency under the time-varying channel state and long-term constraints of off-chain reputation, and solve the problem using Lyapunov optimization method. Experimental results show that the proposed dynamic DNN partitioning and resource allocation (DPRA) algorithm outperforms the baselines in terms of reducing the overall latency while guaranteeing the trustworthiness of the B-DT system.
\end{abstract}


\section{Introduction}
\IEEEPARstart{T}{he} emergence of artificial intelligence (AI) driven Digital Twin (DT) has empowered the industrial Internet of Things (IIoT) to achieve fully industrial automation and smart manufacturing\cite{DBLP:journals/iotj/WuZ021}. With sophisticated interactions between physical entities and their virtual replicas, AI driven DTs enable the performance monitoring, analysis, simulation, and optimization of physical entities to realize predictive maintenance and intelligent decision making\cite{DBLP:journals/tcyb/GuoZRH23}. On the one hand, AI technologies such as deep neural networks (DNNs) improve the accuracy of classification and prediction, and enable effective decision-making processes in DT-assisted IIoT networks~\cite{DBLP:conf/globecom/HuLGZS22}. On the other hand, the state-of-art DNN architectures rely on significant computation resources at the cost of prolonged execution time. Since IIoT networks demand timely completion of each processing step during the manufacturing process, execution latency is a critical performance metric in AI driven DT-assisted IIoT networks\cite{DBLP:journals/tii/LuHZMZ21a,DBLP:journals/tii/GuoTK23}. 

Recent studies focus on communication-efficient DT-assisted IIoT networks. For example, \cite{DBLP:journals/sensors/ChukhnoCACIM20, DBLP:journals/tvt/SunZWZ20, DBLP:journals/jcin/DaiZ22, DBLP:journals/tii/ZhouJLLMGT22, DBLP:journals/iotj/WangLSLMZ23} optimized the communication and computation resource allocation (i.e., transmit power, channel bandwidth, and computation frequency) to minimize the system latency under time-varying channel states and energy consumption constraints. However, lightweight IIoT devices are constrained by insufficient computing power to execute resource-intensive DNN inference tasks. The aforementioned works that mainly concentrate on communication and computation resource allocation are not applicable to AI driven DT-assisted IIoT networks. To address this issue, one approach is to partition the DNN model into two distinct segments and offload the top layers of DNN inference tasks to an edge node with sufficient computing power. Recent works have investigated the applications of DNN partitioning technique to reduce the latency of DNN training and inference in different scenarios\cite{DBLP:conf/msn/WuXL21, DBLP:journals/cn/WangXXJL23, DBLP:journals/jsac/DengLMWSDC23,DBLP:journals/tmc/GaoSSQLL23, DBLP:journals/jcloudc/GuoZS23, DBLP:journals/jstsp/WangCL23}. For example, \cite{DBLP:conf/msn/WuXL21, DBLP:journals/cn/WangXXJL23, DBLP:journals/jsac/DengLMWSDC23} developed different DNN partitioning and offloading strategies to accelerate the training process of federated learning. Later, \cite{DBLP:journals/tmc/GaoSSQLL23, DBLP:journals/jcloudc/GuoZS23, DBLP:journals/jstsp/WangCL23} optimized DNN partitioning point as well as communication and computation resource allocation to reduce the DNN inference latency in mobile edge computing (MEC) networks. However, DNN model partitioning and task offloading can pose a potential risk of malicious attacks from the edge nodes. Additionally, even honest edge nodes may unintentionally transmit confusing or incorrect DNN inference results due to their limited computation capabilities. Therefore, how to design a secure and trustworthy DNN partitioning scheme for AI driven DT-assisted IIoT networks deserves further investigation.

In addition, the reliance of DT technology on analyzing substantial data collected from IIoT devices makes it vulnerable to device malfunctions and cyberattacks. To mitigate these concerns, blockchain as a decentralized distributed and tamper-proof ledger technology has been integrated into DT-assisted IIoT networks to track physical entities and their associated data in a transparent and traceable manner, enabling secure and trustworthy interactions between DTs. For instance, \cite{DBLP:journals/ipm/PutzDEP21, DBLP:journals/iotj/WangCL23, DBLP:journals/tii/LuHZMZ21} proposed different blockchain-enabled DT frameworks for IIoT networks to improve the efficiency and security of data sharing among participants. 

However, computation-intensive consensus algorithms are not affordable for lightweight IIoT devices. Driven by this issue, recent works have proposed different hybrid consensus mechanisms to achieve a trade-off between scalability and security. Authors in \cite{DBLP:journals/iotj/YangYLZL19} and \cite{DBLP:journals/tetc/ZhangLZWK21} proposed different reputation update algorithms and hybrid consensus mechanisms that combine Proof of Work (PoW) and Proof of Stake (PoS) in blockchain-enabled Internet of Vehicles (IoV) networks. Specifically, a roadside unit with a higher reputation is rewarded with a reduced block generation difficulty. In spite of previous effort, how to evaluate, explore, and leverage the endogenous reputation to achieve both reduced block generation latency and system security remains challenging.

To address the above problems, we propose a three-tier blockchain-enabled DT (B-DT) framework for wireless IIoT networks. Specifically, the DTs maintained at the gateway side synchronize the real-time data with the respective IIoT devices, execute the bottom layers of the DNN inference (i.e., bottom-layer DNN inference for short) tasks locally, and offload the top layers of the DNN inference (i.e., top-layer DNN inference for short) to the access point (AP) side. The gateways and APs collaborate to perform the DTs’ DNN inference tasks off the chain, and the APs serve as blockchain nodes to record the inference results on the chain. Our main contributions can be summaried as follows.
\begin{itemize}
	\item We propose to partition DTs' DNN inference tasks between the gateway side and the AP side, where the gateways execute the bottom-layer DNN inference tasks, and offload the top-layer DNN inference tasks to the AP side. Owing to the sufficient computing power of the APs, the DNN model partitioning and task offloading scheme achieves a significant reduction in DNN inference latency.  
	\item We design a novel reputation-based consensus mechanism that adaptively tunes the block generation difficulty. At its core, the proposed reputation-based consensus mechanism evaluates the off-chain reputation of each AP according to its computation resource contributions to the DNN inference tasks, and utilizes the off-chain reputation as a stake to adjust the block generation difficulty of the on-chain PoW. In this way, the degradation of PoW consensus security is compensated by off-chain reputation stake, which improves the consensus efficiency while ensuring the trustworthiness on the chain.
	\item To obtain a communication and computation efficient wireless B-DT system, we formulate a joint dynamic optimization problem of communication resource (i.e., partition point) and computation resource allocation (i.e., computation frequency of APs for top-layer DNN inference and block generation) under the time-varying wireless channel state and the constraints of energy consumption and off-chain reputation. The long-term off-chain reputation constraint is adopted to bound the average off-chain reputation of each AP to ensure both scalability and trustworthiness of the B-DT system.
	\item We analyze the performance of the proposed dynamic \textbf{D}NN \textbf{p}artitioning and \textbf{r}esource \textbf{a}llocation (DPRA) algorithm in terms of asymptotic optimality, and characterize an [$\mathcal{O}(1/V)$, $\mathcal{O}(V)$] trade-off between the minimization of system latency and the satisfaction of the long-term off-chain reputation constraint with a control parameter $V$. This trade-off indicates that the minimizing system latency and guaranteeing system trustworthiness can be balanced by adjusting $V$. 
	\item Experimental results demonstrate the performance of DPRA algorithm in terms of system latency and off-chain reputation, and show that DPRA outperforms the baselines in terms of reducing the overall latency while guaranteeing the trustworthiness of the B-DT system.
\end{itemize}

We organize this paper as follows. Section \ref{Preliminaries} introduces the concepts of DT, blockchain, and DNN partitioning. In Section \ref{System Model}, we first present the blockchain-enabled DT framework for wireless IIoT networks, and then formulate the stochastic optimization problem. In Section \ref{Problem Solution}, we present the DPRA algorithm. Section \ref{Performance Analysis} investigates the trade-off between minimizing system latency and satisfying the long-term off-chain reputation constraint. The experimental results are then shown in Section \ref{Experimental Results}. Section \ref{Conclusion} concludes the paper.

\section{Preliminaries}\label{Preliminaries}
\subsection{Digital Twin}
A digital twin (DT) refers to a virtual representation of a physical entity. The concept of DT consists of two distinct tiers: physical twin tier and digital twin tier. In the physical twin tier, entities synchronize real-time data and state information captured by equipped sensors and running applications with the digital twin tier. Meanwhile, the digital twin tier utilizes the collected data and simulation models to generate virtual entities,  which facilitates the prediction of future scenarios in the real world. Ultimately, the optimal decisions derived from these predictions are fed back to the physical entities to enhance their performance. 

By employing data analytics, machine learning (ML), and AI technologies, DTs can serve multiple purposes in IIoT networks such as monitoring, analysis, prediction, and optimization, which enables IIoT devices to achieve predictive maintenance, make well-informed decisions, and improve operational efficiency\cite{DBLP:journals/iotj/LuHZMZ21}.	

\subsection{Blockchain}
Blockchain is a secure, transparent, and decentralized distributed ledger technology. The chain structure ensures the immutability of transaction records by linking each block with a unique hash to the previous one. Moreover, the consensus mechanism enhances the data security and integrity by facilitating agreement among all participating nodes regarding the validity of transactions. Blockchain technology has been introduced into DT to facilitate decentralized data storage and immutable record keeping. This integration drives innovation in IIoT networks by providing a reliable and trustworthy infrastructure for managing and sharing data\cite{DBLP:journals/network/YaqoobSUJOI20}.

As a vital component of blockchain technology, a distributed consensus mechanism is essential to achieve distributed consistency on the data among trustless nodes in a decentralized manner. In practice, Proof of Work (PoW)\cite{nakamoto2008bitcoin} and Proof of Stake (PoS)\cite{shi2022pooling} are the two most popular consensus mechanisms in blockchain. In PoW, each blockchain node competes to find a nonce number that matches the hash value of the transactions to a predetermined target. Once a node successfully finds the target nonce, it is granted the privilege to publish a new block. However, solving this cryptographic hash puzzle consumes significant computation resources for the participating nodes\cite{DBLP:journals/tpds/LiSWDMSHP22}. PoS, on the other hand, is based on the concept of ``staking". Instead of computational power consumption, nodes participate in the PoS consensus process by providing verifiable stakes in the cryptocurrency. The probability of being selected to validate transactions and publish new blocks is directly proportional to the number of stakes held by the blockchain node. Compared to PoW, PoS consumes much less energy and thereby reduces the need for expensive mining hardware. However, the lack of significant participation thresholds exposes it to potential long-range\cite{DBLP:journals/tsc/FengMMLC23} and sybil attacks\cite{DBLP:journals/cn/PlattM21}, which ultimately leads to diminished decentralization and compromised security.	
\begin{figure*}[!t]
	\centering
	\includegraphics[width=6.7in]{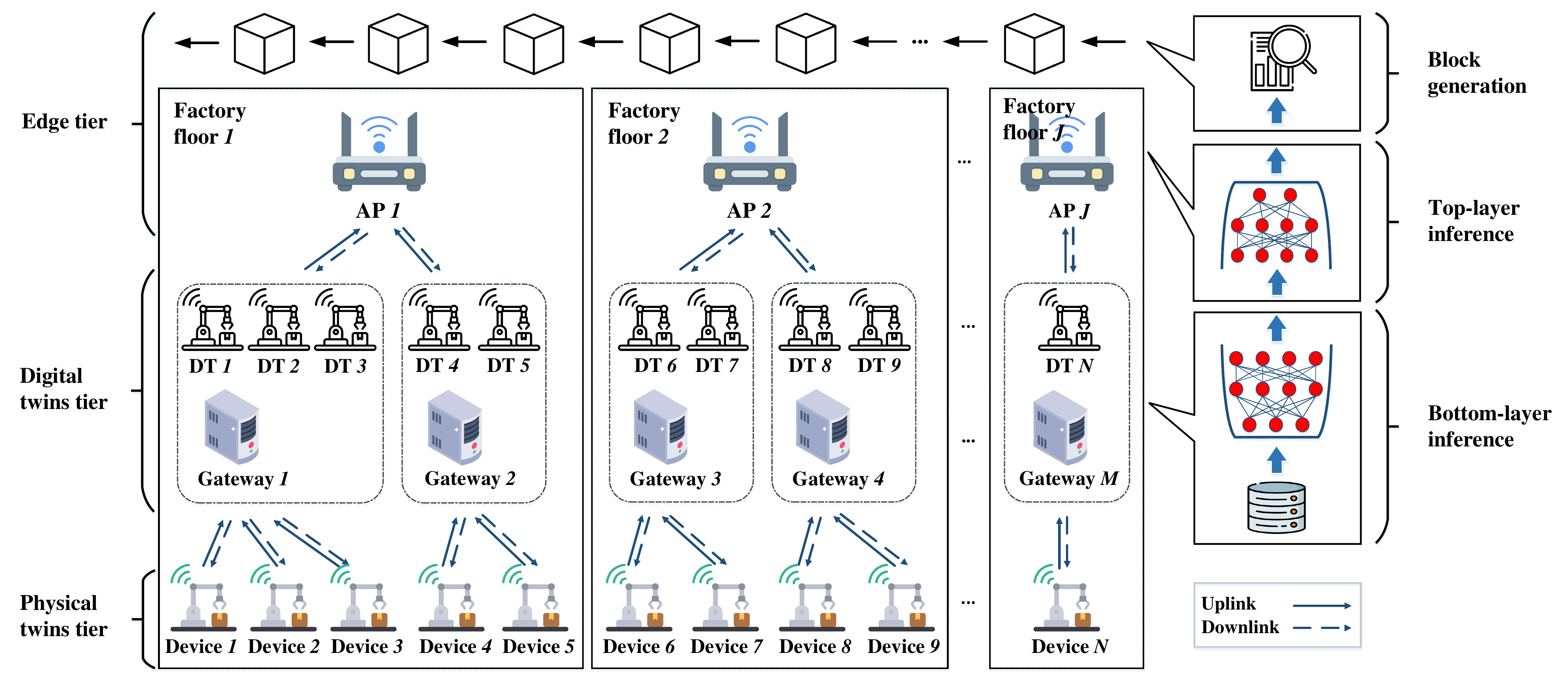}
	\caption{The system model of the three-tier blockchain-enabled DT framework for wireless IIoT networks.}
	\label{fig: system model}
\end{figure*}
\subsection{Deep Neural Network Partitioning}
In the process of DNN inference, raw data is sequentially passed through multiple hidden layers and subjected to various mathematical operations and transformations\cite{DBLP:journals/ton/ZengCZYZ21}. Specifically, the input undergoes weighted and biased, applies non-linear activation functions at each layer, and then passes to the subsequent layer until it reaches the output layer.

DNN partitioning divides a DNN model into multiple segments, which facilitates the deployment and execution of DNN models across different computing platforms\cite{DBLP:journals/jsac/DengLMWSDC23}. Partitioning a DNN model involves selecting a specific layer within the network (referred to as the partition point) to separate the model into two distinct segments. The bottom layer refers to the hidden layer that extends from the input layer to the partition point, whereas the top layer represents the hidden layer spanning from the partition point to the output layer. By employing DNN partitioning, DTs can offload top-layer DNN inference tasks to an edge server, leading to a significant reduction in both computation load and execution latency.
\section{System Model}\label{System Model}
In this paper, we consider a blockchain-enabled DT (B-DT) system in wireless IIoT networks. As shown in Fig.\ref{fig: system model}, the B-DT system consists of three tiers: the physical twin tier with multiple IIoT devices, the digital twin tier with multiple gateways, and the edge server tier with multiple access points (APs). Each device holds a local dataset that is continuously collected from its equipped sensors and running applications, and keeps synchronising its collected data with the corresponding DT maintained on its associated gateway. Let $\mathcal{T}=\{0,1,..., T\}$ denote the time index set. In each time slot, the proposed B-DT system operates in the following steps:\begin{enumerate}
	\item \textit{DNN inference:} At the beginning of each time slot, each gateway runs its locally maintained DTs to perform the bottom-layer DNN inference, and transmits the forward output of the bottom-layer DNN inference to its associated AP. Upon receiving the forward output of the bottom-layer DNN inference from the gateways, each AP performs the top layers of the DNN inference, and finally outputs its inference results. 
	\item \textit{Block mining:} Each AP encrypts its DNN inference results by a unique digital signature, and exchanges the DNN inference results with the other APs over the peer-to-peer network. Then, the APs add the verified inference results to their respective candidate block, and compete to generate a new block with the proposed consensus mechanism.
\end{enumerate}

\subsection{DNN Inference and Offloading}
Denote the index sets of the devices, gateways and APs by $\mathcal{N}=\{1,2,...,N\}$, $\mathcal{M}=\{1,2,...,M\}$, and $\mathcal{J}=\{1,2,...,J\}$, respectively. Define an $N\times M$ connection matrix as $\boldsymbol{a}$ with entry $a_{n,m}\in\{0,1\}$, $\forall n\in\mathcal{N}$, and $m\in\mathcal{M}$. If $a_{n,m}=1$, the $n$-th device is deployed with the $m$-th gateway on the same factory floor, and communicates with the $m$-th gateway to maintain its corresponding DT. We refer to the corresponding DT of the $n$-th device as the $n$-th DT. Note that the corresponding DT of each device is maintained by a single gateway, i.e., $\sum_{m\in\mathcal{M}}a_{n,m}=1$, $\forall n\in\mathcal{N}$. From \cite{DBLP:journals/tpds/YangBYTZ22}, we formulate the stochastic data arrivals at the gateway side as a homogeneous Poisson process. To be specific, the new data points $D_n(t)$ that collected by the $n$-th device and transmitted to its corresponding DT in each time slot is an independent and identically distributed (i.i.d.) exponential random variable with an average rate $\Theta_n$. Note that the average rate $\Theta_n$ is positively correlated with both the data collection rate of the $n$-th device and the data transmission rate from the $n$-th device to the associated gateway.

Before calculating the latency and energy consumption for the DNN inference at the gateway and AP sides, we first introduce the notations for hyper-parameters and tensor shapes involved in the DNN inference process as follows. Let $B_s$ and $S_f$ denote the batch size and the precision format of the data type, respectively. For convolution layers and pooling layers, $H_o$, $W_o$ and $C_o$ represent the output height, width, and channel, respectively; $H_i$, $W_i$ and $C_i$ represent the input height, width, and channel; $H_f$ and $W_f$ are the height and width of the filter. For the fully connected layers, $S_i$ and $S_o$ refer to the input and output sizes, respectively. We present the main layer-level forward output size and floating-point operation counts (FLOPs) in the DNN inference process in Table \ref{table1} to calculate the number of bits of the forward output from the bottom-layer DNN inference and FLOPs for the bottom-layer and top-layer DNN inference performed at the gateway and AP sides.\begin{table}[!t]
	\caption{Layer-level forward output size and FLOPs in the DNN inference process.}
	\centering\label{table1}
	\renewcommand{\arraystretch}{1.2}
	\begin{tabular}{!{\vrule width 1pt}c|c|c!{\vrule width 1pt}}
		\noalign{\hrule height 1pt}
		\textbf{Layer Category}& \textbf{Forward Output Size} & \textbf{FLOPs}\\
		\noalign{\hrule height 1pt}
		Convolution &$S_f B_s C_o H_o W_o$ & $2B_s C_iH_fW_fC_o H_o W_o$\\
		\noalign{\hrule height 1pt}
		Pooling &$S_f B_s C_o H_o W_o$ &$B_s C_iH_iW_i$\\
		\noalign{\hrule height 1pt}
		Fully Connected &$B_s S_o$ & $2 B_s S_i S_o$\\
		\noalign{\hrule height 1pt}
	\end{tabular}
\end{table}

Let $\mathcal{L}_n=\{1,...,L_n\}$ denote the index set of the DNN layers for the $n$-th DT. In the $t$-th time slot, the bottom $l_n(t)$ layers of the DNN inference are executed locally at the gateway side, and the top $L_n-l_n(t)$ layers of the DNN inference tasks are offloaded to the AP side. Let $\chi_n^l$ denote the FLOPs required by the $n$-th DT to perform the $l$-th layer of the DNN inference for each data point. As such, the DNN inference time of the $m$-th gateway in the $t$-th time slot is\begin{alignat}{1}
	\tau_m^\text{exe,G}(t)=\frac{\sum_{n\in\mathcal{N}}a_{n,m}D_n(t)\sum_{l=1}^{l_n(t)}\chi_n^l}{\phi_m^\text{G} f_m^\text{G}},
\end{alignat}where $\phi_m^\text{G}$ is the FLOPs per clock cycle of the $m$-th gateway, and $f_m^\text{G}$ is the computation frequency of the $m$-th gateway for the DNN inference. The energy consumption of the $m$-th gateway for DNN inference tasks in the $t$-th time slot is
\begin{alignat}{1}
	e_m^\text{exe,G}(t)=\frac{v_m^\text{G}\left(f_m^\text{G}\right)^2}{\phi_m^\text{G}}\sum_{n\in\mathcal{N}}a_{n,m}D_n(t)\sum_{l=1}^{l_n(t)}\chi_n^l,
\end{alignat}where $v_m^\text{G}$ is the effective switched capacitance. 

Assume that the wireless channels between the gateways and APs experience i.i.d. block fading. Specifically, the wireless channel remains static in each time slot but varies among different time slots. We model the data transmission between the gateways and APs using the multiple channel access method of orthogonal frequency division multiple access (OFDMA). Define an $M\times J$ connection matrix as $\boldsymbol{b}$ with entry $b_{m,j}\in\{0,1\}$, $\forall m\in\mathcal{M}$, and $j\in\mathcal{J}$. If $b_{m,j}=1$, the $m$-th gateway is deployed with the $j$-th AP on the same factory floor, and offloads the top-layer DNN inference tasks to the $j$-th AP via wireless link in each time slot. Notably, each gateway can only communicate with the AP deployed on the same shop floor, i.e., $\sum_{j\in\mathcal{J}}b_{m,j}=1$, $\forall m\in\mathcal{M}$. The uplink channel power gain from the $m$-th gateway to its associated AP is modeled as\begin{alignat}{1}
	H_m(t)=h_0\rho_m(t)\left({d_0}/{d_m}\right)^\nu,
\end{alignat}where $h_0$ is the path loss constant, $\rho_m(t)$ is the small-scale fading channel power gain from the $m$-th gateway to its associated AP in the $t$-th time slot, $d_m$ is the distance from the $m$-th gateway to its associated AP, $d_0$ is the reference distance, and $\nu$ is the large-scale path loss factor, respectively. Let $o_n^l$ denote the forward output size of the $l$-th layer for the $n$-th DT in the DNN inference process. The data transmission time from the $m$-th gateway to the associated AP is\begin{alignat}{1}
	\tau_m^\text{off}(t)=\frac{\sum_{n\in\mathcal{N}}a_{n,m}D_n(t)o_n^{l_n(t)}}{B\log\left(1+\frac{P_mH_m(t)}{\eta_m(t)+N_0B}\right)},
\end{alignat}where $P_m$ is the transmit power of the $m$-th gateway, $N_0$ is the noise power spectral density, and $\eta_m(t)$ is the co-channel interference. The energy consumption of the $m$-th gateway for DNN inference task offloading is represented as\begin{alignat}{1}
e_m^\text{off}(t)=\frac{P_m\sum_{n\in\mathcal{N}}a_{n,m}D_n(t)o_n^{l_n(t)}}{B\log\left(1+\frac{P_mH_m(t)}{\eta_m(t)+N_0B}\right)}.
\end{alignat}

The time consumption for the top-layer DNN inference offloaded from the $m$-th gateway is represented as\begin{alignat}{1}
	\tau_m^\text{exe,A}(t)=\frac{\sum_{n\in\mathcal{N}}a_{n,m}D_n(t)\sum_{l=l_n(t)+1}^{L_n}\chi_n^l}{\phi_m^\text{A} f_m^\text{A}(t)},
\end{alignat}where $\phi_m^\text{A}$ is the FLOPs per clock cycle of the $m$-th gateway's associated AP, and $f_m^\text{A}(t)$ is the computation frequency of the $m$-th gateway's associated AP for DNN inference tasks. The energy consumption for the top-layer DNN inference offloaded from the $m$-th gateway is represented as\begin{alignat}{1}
e_m^\text{exe,A}(t)=\frac{v_m^\text{A}\left(f_m^\text{A}(t)\right)^2}{\phi_m^\text{A}}\sum_{n\in\mathcal{N}}a_{n,m}D_n(t)\sum_{l=l_n(t)+1}^{L_n}\chi_n^l,
\end{alignat}where $v_m^\text{A}$ is the effective switched capacitance of the $m$-th gateway's associated AP.

\subsection{Reputation Based Consensus Mechanism}
In this subsection, we develop a reputation-based hybrid consensus mechanism by integrating PoS and PoW. Before delving into the reputation-based consensus mechanism, we first design an off-chain reputation evaluation mechanism to evaluate the off-chain reputation of each AP as follows. 

\textit{Off-chain reputation evaluation:} Calculating the top-layer DNN reference tasks offloaded to each AP, we evaluate the off-chain reputation of the $j$-th AP in the $t$-th time slot as\begin{alignat}{1}\label{off-chain}
	U_j(t)= g\left(O_j(t)\right),
\end{alignat}where $g(\cdot)$ is a generic off-chain reputation evaluation function, which can be further customized according to specific reputation evaluation rules in diverse networks, and
\begin{alignat}{1}
	O_j(t)=\sum_{m\in\mathcal{M}}\sum_{n\in\mathcal{N}}b_{m,j}a_{n,m}D_n(t)\sum_{l=l_n(t)+1}^{L_n}\chi_n^l\end{alignat}is the top-layer DNN reference tasks offloaded from the associated gateways to the $j$-th AP. From (\ref{off-chain}), the AP with more computational contributions to the DNN inference tasks can achieve higher off-chain reputation values.

The core of the proposed reputation-based consensus mechanism is that the off-chain reputation can be used as the stake to adjust the block generation difficulty of on-chain PoW. In this way, the degradation of PoW consensus security is compensated by off-chain reputation stake, which improves the consensus efficiency while ensuring the trustworthiness on the chain.

\textit{Block generation difficulty:} The block generation difficulty is inversely proportional to the off-chain reputation of the APs. From \cite{DBLP:journals/iotj/YangYLZL19}, the relationship between the block generation difficulty and off-chain reputation is represented as\begin{alignat}{1}
	\gamma_j(t)=e^{-\alpha U_j(t)-\beta},
\end{alignat}where $\alpha$ and $\beta$ control the influence of the off-chain reputation values on block generation difficulty and the final convergence of block generation difficulty, respectively. 

We next clarify the statistical property of the stochastic process with respect to the block generation. From \cite{DBLP:journals/tcom/PokhrelC20,DBLP:journals/iotj/QuGLXYLZ20}, the successful query attempts of the $j$-th AP converge to a Poisson process with the average rate given by\begin{alignat}{1}
	\theta_j(t)=\frac{f_j^\text{bloc}(t)}{\gamma_j(t)},\end{alignat}where $f_j^\text{bloc}(t)$ is the computation frequency of the $j$-th AP for block generation in the $t$-th time slot. Since the APs work independently on the hash puzzle in each time slot, the successful query attempts of the entire blockchain network can be formulated as a Poisson process with the average rate\cite{prokhorov1983probability}:\begin{alignat}{1}
	\hat{\theta}(t)=\sum_{j\in\mathcal{J}}\theta_j(t).\end{alignat}As such, the block generation time in each time slot $\tau^\text{bloc}(t)$ , denoted by $\tau^\text{bloc}(t)$, can be formulated as an i.i.d. exponential random variable with the average rate $\hat{\theta}(t)$. Therefore, the cumulative distribution function of the block generation time in the $t$-th time slot is given by\begin{alignat}{1}
	\text{Pr}(\tau^\text{bloc}(t)<\tau)=1-e^{-\hat{\theta}(t)\tau}.\end{alignat} Assume that the new block can be generated when $p_0=\text{Pr}(\tau^\text{bloc}<\tau)$ approaches one. Thus, the block generation time in the $t$-th time slot is given by\begin{equation}
\tau^\text{bloc}(t)=-\frac{\ln(1-p_0)}{\hat{\theta}(t)}.
\end{equation}The energy consumption of the $j$-th AP for block generation in the $t$-th time slot is expressed as \begin{equation}
e_j^\text{bloc}(t)=v_j^A\tau^\text{bloc}(t){\left(f_j^\text{bloc}(t)\right)}^3.
\end{equation}

\subsection{Problem Formulation}
According to the analysis above, the total latency of the wireless B-DT system mainly comes from four parts, i.e., bottom-layer DNN inference at the gateway side, data transmission from the gateway side to the AP side, top-layer DNN inference at the AP side, and block generation. Thus, the system latency in each time slot is given by\begin{equation}
	\tau(t)=\max_{m\in\mathcal{M}}\left\{\tau_m^\text{exe,G}(t)+\tau_m^\text{off}(t)+\tau_m^\text{exe,A}(t)\right\}+\tau^\text{bloc}(t).
\end{equation}The total energy consumption of the $m$-th gateway in the $t$-th time slot is\begin{equation}
e_m^\text{G}(t)=e_m^\text{exe,G}(t)+e_m^\text{off}(t).
\end{equation}The energy consumption of the \!$j$-th AP in the $\!t$-th time slot is\begin{equation}
e_j^\text{A}(t)=\sum_{m\in\mathcal{M}}b_{m,j}e_m^\text{exe,A}(t)+e_j^\text{bloc}(t).
\end{equation}Let $E_m^\text{\rm{G}}(t)$ and $E_j^\text{\rm{A}}(t)$ denote the energy arrivals at the $m$-th gateway and the $j$-th AP in the $t$-th time slot, respectively. Consider that $E_m^\text{\rm{G}}(t)$ and $E_j^\text{\rm{A}}(t)$ are modeled as i.i.d. stochastic processes, and are uniformly distributed within $[0, E_m^\text{\rm{G},max}]$ and $[0, E_j^\text{\rm{A},max}]$. Notably, the energy consumption of the gateways and APs in each time slot cannot exceed the respective available energy, i.e., $0\leq e_m^\text{\rm{G}}(t)\leq E_m^\text{\rm{G}}(t)$, and $0\leq e_j^\text{\rm{A}}(t)\leq E_j^\text{\rm{A}}(t)$.

To obtain a communication and computation efficient wireless B-DT system, we jointly optimize the communication resource (i.e., DNN partition point) and the computation resource (i.e., computation frequency for DNN inference and block generation) allocation. Let $\boldsymbol{X}(t)\!=\![\boldsymbol{l}(t), \boldsymbol{f}^\text{A}(t), \boldsymbol{f}^\text{bloc}(t)]$. The stochastic optimization problem is formulated as\begin{alignat}{1}\label{P}
&\quad\textbf{P0}:\;\min_{\boldsymbol{X}(t)}\; \overline{\tau}=\frac{1}{T}\sum_{t=1}^T\tau(t)\\
\text{s.t.}\quad 
&\textbf{C1}: 1\leq l_n(t)\leq L_n,\forall n\in \mathcal{N},t\in \mathcal{T},\nonumber\\
&\textbf{C2}: 0\leq{\sum}_{m\in\mathcal{M}}b_{m,j}f_m^\text{\rm{A}}(t)\leq f_j^\text{\rm{max}}, \forall j\in \mathcal{J}, t\in \mathcal{T},\nonumber\\
&\textbf{C3}: 0\leq f_j^\text{\rm{bloc}}(t)\leq f_j^\text{\rm{max}}, \forall j\in \mathcal{J}, t\in \mathcal{T},\nonumber\\
&\textbf{C4}: 0\leq e_m^\text{\rm{G}}(t)\leq E_m^\text{\rm{G}}(t), \forall m\in \mathcal{M}, t\in \mathcal{T},\nonumber\\
&\textbf{C5}: 0\leq e_j^\text{\rm{A}}(t)\leq E_j^\text{\rm{A}}(t), \forall j\in \mathcal{J}, t\in \mathcal{T},\nonumber\\
&\textbf{C6}: U^\text{\rm{min}}\leq\frac{1}{T}\sum_{t=1}^TU_j(t)\leq U^\text{\rm{max}}, \forall j\in \mathcal{J},\nonumber
\end{alignat}where the constraints \textbf{C1}$\sim$\textbf{C3} bound the ranges of the variables $\boldsymbol{l}(t)$, $\boldsymbol{f}^\text{A}(t)$, and $\boldsymbol{f}^\text{bloc}(t)$, respectively. \textbf{C4} and \textbf{C5} are the energy consumption constraints for devices and gateways in each time slot, respectively. In order to guarantee both scalability and trustworthiness of the B-DT system, the long-term constraint \textbf{C6} is adopted to bound the average off-chain reputation of each AP. Specifically, the lower bound of the average off-chain reputation (i.e., $\frac{1}{T}\sum_{t=1}^TU_j(t)\ge U^\text{\rm{min}}$) imposes an upper bound on the block generation difficulty, and thereby ensuring a low block generation latency. The upper bound of the average off-chain reputation (i.e., $\frac{1}{T}\sum_{t=1}^TU_j(t)\leq U^\text{\rm{max}}$) imposes a lower bound on the block generation difficulty, which guarantees the trustworthiness of the B-DT system.  

Therefore, our goal is to minimize the total latency of the blockchain-enabled DT system with the time-varying wireless channel state and the constraints of energy consumption and off-chain reputation.

\section{Problem Solution}\label{Problem Solution}
In this section, we introduce a dynamic \textbf{D}NN \textbf{p}artitioning and \textbf{r}esource \textbf{a}llocation (DPRA) algorithm to solve the long-term stochastic problem in \textbf{P0}, which is shown in \textbf{Algorithm} 1. By leveraging the Lyapunov optimization method, DPRA first transforms the long-term stochastic optimization problem in \textbf{P0} into a sequence of one-shot static optimization problems, and then subsequently solves the transformed deterministic problem in each time slot. Unlike existing DNN partitioning approaches that employ a pre-defined partition point, the proposed DPRA algorithm dynamically optimizes both the DNN partitioning point and the computation frequency considering time-varying channels and energy arrivals.

\begin{algorithm}[!t]
	\caption{Dynamic DNN partitioning and resource allocation (DPRA) algorithm}\label{alg::Dynamic Resource Allocation and Client Scheduling Algorithm}
	Initialize:	Auxiliary queue lengths $\boldsymbol{Q}(t)=\frac{1}{2}\left(U^\text{min}+U^\text{max}\right)$, $\boldsymbol{S}(t)=\frac{1}{2}\left(U^\text{min}+U^\text{max}\right)$;\\
	\For{$t=1,2,...,T$}{
		\algorithmicrequire{ Auxiliary queue lengths and channel state in the $t$-th time slot};\\
		\algorithmicensure{ $\boldsymbol{X}(t)=\left[\boldsymbol{l}(t), \boldsymbol{f}^\text{A}(t), \boldsymbol{f}^\text{bloc}(t)\right]$};\\
		\DontPrintSemicolon
		\SetKwBlock{DoParallel}{do in parallel}{end}
		\DoParallel{
			Optimize the DNN partirion point $\boldsymbol{l}(t)$, the computation frequency for the top-layer DNN inference $\boldsymbol{f}^\text{A}(t)$, and the computation frequency for block generation $\boldsymbol{f}^\text{bloc}(t)$ by solving (\ref{P2_1}), (\ref{P2_2}), and (\ref{P2_3}) with block coordinate descent method;
		}
		Update $\boldsymbol{Q}(t)$ and $\boldsymbol{S}(t)$ according to (\ref{q1}) and (\ref{q2});\\
		\algorithmicreturn{ $\boldsymbol{X}(t)=\left[\boldsymbol{l}(t), \boldsymbol{f}^\text{A}(t), \boldsymbol{f}^\text{bloc}(t)\right]$}
	}
\end{algorithm}

\subsection{Problem Transformation}
To decouple the long-term stochastic optimization problem presented  in \textbf{P0} into a sequence of one-shot static problems, we first define two auxiliary queues for each AP as\begin{equation}\label{q1}
	Q_j(t+1)=\max\left\{Q_j(t)-U_j(t)+U^\text{min},0\right\},
\end{equation}and\begin{equation}\label{q2}
	S_j(t+1)=\max\left\{S_j(t)+U_j(t)-U^\text{max},0\right\}.
\end{equation}According to Lyapunov optimization method\cite{DBLP:journals/twc/DengLMWSDCP23,DBLP:journals/tmc/DengLSWZY22,DBLP:series/synthesis/2010Neely}, the long-term off-chain reputation constraint \textbf{C6} can be equivalently transformed into the queue stability constraints for the auxiliary queues $Q_j(t)$ and $S_j(t)$ in (\ref{q1}) and (\ref{q2}) as\begin{equation}\label{s1}
\lim_{t\to \infty}\frac{\mathbb{E}\left\{\left\vert Q_j(t)\right\vert\right\}}{t}=0, \forall j\in\mathcal{J},
\end{equation}and\begin{equation}\label{s2}
\lim_{t\to \infty}\frac{\mathbb{E}\left\{\left\vert S_j(t)\right\vert\right\}}{t}=0, \forall j\in\mathcal{J}.
\end{equation}By substituting the long-term inequality constraint \textbf{C6} with the queue stability constraints in (\ref{s1}) and (\ref{s2}), the original problem in \textbf{P0} can be rewritten as\begin{alignat}{1}\label{P1}
\quad&\textbf{P1}:\;\min_{\boldsymbol{X}(t)}\; \overline{\tau}=\frac{1}{T}\sum_{t=1}^T\tau(t)\\
\text{s.t.}\quad 
&\textbf{C1}\sim\textbf{C5},\nonumber\\
&\textbf{C7}: \lim_{t\to \infty}\frac{\mathbb{E}\left\{\left\vert Q_j(t)\right\vert\right\}}{t}=0, \forall j\in\mathcal{J},\nonumber\\
&\textbf{C8}: \lim_{t\to \infty}\frac{\mathbb{E}\left\{\left\vert S_j(t)\right\vert\right\}}{t}=0, \forall j\in\mathcal{J}.\nonumber
\end{alignat}

The transformed problem \textbf{P1} is now a standard optimization problem for the Lyapunov optimization method. To solve \textbf{P1}, we proceed by formulating the Lyapunov function, characterizing the conditional Lyapunov drift, and minimizing the Lyapunov drift-plus-penalty ratio function as follows.

\textit{Lyapunov function}: The Lyapunov function is defined as\begin{equation}\label{Lyapunov function}
	L(t)=\frac{1}{2}\sum_{j\in\mathcal{J}}\left(\left(Q_j(t)\right)^2+\left(S_j(t)\right)^2\right).
\end{equation} 

\textit{Conditional Lyapunov drift}: Let $\boldsymbol{\Phi}(t)=\big\{Q_j(t), S_j(t), \forall j\in\mathcal{J}\big\}$ be a set that collects all auxiliary queues in the $t$-th time slot. We define the conditional Lyapunov drift as\begin{equation}
	\Delta L(t)=\mathbb{E}\left\{L(t+1)-L(t)\vert\boldsymbol{\Phi}(t)\right\}.
\end{equation}

\textit{Lyapunov drift-plus-penalty function}: Given any Lyapunov control parameter $V>0$, we define the Lyapunov drift-plus-penalty function as\begin{equation}
	\Delta (t)=V\tau(t)+\Delta L(t).
\end{equation}From \cite{DBLP:journals/twc/DengLMWSDCP23,DBLP:journals/tmc/DengLSWZY22,DBLP:series/synthesis/2010Neely}, minimizing the conditional Lyapunov drift $\Delta L(t)$ promotes the stability of the auxiliary queues $Q_j(t)$ and $S_j(t)$. Therefore, we minimize the Lyapunov drift-plus-penalty function to reduce the system latency while ensuring the stability of the auxiliary queues. Note that the control parameter $V$ allows us to adjust the trade-off between minimizing system latency and satisfying the auxiliary queue stability constraints \textbf{C7} and \textbf{C8}.

\begin{lemma}\label{lemma1}
Given the auxiliary queue lengths $\boldsymbol{\Phi}(t)$, the conditional Lyapunov drift $\Delta L(t)$ is upper bounded by\begin{alignat}{1}\label{euqlemma1}
	\Delta L(t)\leq &\sum_{j\in\mathcal{J}}\mathbb{E}\left\{Q_j(t)\left(U^\text{min}-U_j(t)\right)\right.\nonumber\\&\left.\left.+S_j(t)\Big(U_j(t)-U^\text{max}\Big)\right\vert\boldsymbol{\Phi}(t)\right\}+H,
\end{alignat}where $H=\sum_{j\in \mathcal{J}} g\left(\sum_{m\in\mathcal{M}}\sum_{n\in\mathcal{N}}b_{m,j}a_{n,m}D_n(t)\sum_{l=1}^{L_n}\chi_n^l\right)^2+\frac{J}{2}\left(\left(U^\text{min}\right)^2+\left(U^\text{max}\right)^2\right)$.
\end{lemma}\begin{IEEEproof}
Please see Appendix A.
\end{IEEEproof}Based on the derived upper bound of the conditional Lyapunov drift $\Delta L(t)$ in (\ref{euqlemma1}), the goal of the DPRA algorithm is to minimize the Lyapunov drift-plus-penalty function in each time slot as\begin{alignat}{1}\label{P2}
	\quad&\textbf{P2}:\;\min_{\boldsymbol{X}(t)}\; V\tau(t)+\sum_{j\in\mathcal{J}}\left(S_j(t)-Q_j(t)\right)U_j(t)\\
	\text{s.t.}\quad 
	&\textbf{C1}\sim\textbf{C5}.\nonumber
\end{alignat}
\subsection{Solution of \textbf{P2}}
In this subsection, we resolve the mixed-integer nonlinear problem in \textbf{P2} using the block coordinate descent method. As shown in \textbf{Algorithm} 1, we first decompose (\ref{P2}) into three sub-problems in (\ref{P2_1}), (\ref{P2_2}), and (\ref{P2_3}), and solve the sub-problems in a successive way.

Fixing the DNN partitioning point $\boldsymbol{l}(t)$ and the computation frequency $\boldsymbol{f}^\text{A}(t)$ for the top-layer DNN reference, the problem in \textbf{P2} can be rewritten as\begin{alignat}{1}\label{P2_1}
	\min_{\boldsymbol{f}^\text{bloc}(t)}\;&g_1\left(\boldsymbol{f}^\text{bloc}(t)\right)=\frac{-\ln(1-p_0)}{\sum_{j\in\mathcal{J}}f_j^\text{bloc}(t)e^{\alpha U_j(t)+\beta}}\\
	\text{s.t.}\quad 
	&\textbf{C3}, \textbf{C5}.\nonumber
\end{alignat}Let $g_1^*(t)$ denote the minimum value of the objective function in (\ref{P2_1}). Let $g_1^\text{min}(t)=\frac{-\ln(1-p_0)}{\sum_{j\in\mathcal{J}}f_j^\text{max}e^{\alpha U_j(t)+\beta}}$ and $g_1^\text{max}(t)=\frac{-\ln(1-p_0)}{\sum_{j\in\mathcal{J}}f_j^\text{min}e^{\alpha U_j(t)+\beta}}$ denote the lower and upper bound of $g_1^*(t)$. Using the bisection method, we solve the sub-problem in (\ref{P2_1}) by iteratively narrowing the interval between the lower and upper bound of the minimum value of the objective function in (\ref{P2_1}), i.e., $\left[g_1^\text{min}(t), g_1^\text{max}(t)\right]$. In each iteration, we first calculate the mid point of the interval $\left[g_1^\text{min}(t), g_1^\text{max}(t)\right]$, i.e., $\mu=\frac{1}{2}\left(g_1^\text{min}(t)+g_1^\text{max}(t)\right)$, and then compute the upper bound of $f_j^\text{bloc}(t)$ according to the constraints \textbf{C3} and \textbf{C5}, i.e., $f_j^\text{bloc}(t)\leq \min\left\{f_j^\text{max}, \sqrt[3]{\frac{E_j^\text{\rm{A}}(t)-\sum_{m\in\mathcal{M}}b_{m,j}e_m^\text{exe,A}(t)}{v_j^A\mu}}\right\}$. If the mid point $\mu>\frac{-\ln(1-p_0)}{\sum_{j\in\mathcal{J}}\min\left\{f_j^\text{max}, \sqrt[3]{\frac{E_j^\text{\rm{A}}(t)-\sum_{m\in\mathcal{M}}b_{m,j}e_m^\text{exe,A}(t)}{v_j^A\mu}}\right\}e^{\alpha U_j(t)+\beta}}$, the interval between the lower and upper bound of the minimum value of the objective function in (\ref{P2_1}) can be narrowed down by updating $g_1^\text{max}(t)=\mu$. Otherwise, the lower bound of $g_1^*(t)$ is updated by $g_1^\text{min}(t)=\mu$. Thus, the optimal computation frequency for block generation is derived as\begin{small}\begin{alignat}{1}
\!f_j^\text{bloc}(t)=\min\left\{f_j^\text{max}, \sqrt[3]{\frac{E_j^\text{\rm{A}}(t)-\sum_{m\in\mathcal{M}}b_{m,j}e_m^\text{exe,A}(t)}{v_j^A\mu}}\right\}.\!\!
\end{alignat}\end{small}

Fixing the DNN partitioning point $\boldsymbol{L}(t)$ and the computation frequency $\boldsymbol{f}^\text{bloc}(t)$ for block generation, the problem in \textbf{P2} can be rewritten as\begin{alignat}{1}\label{P2_2}
	&\min_{\boldsymbol{f}^\text{A}(t)}\;g_2\left(\boldsymbol{f}^\text{A}(t)\right)=\max_{m\in\mathcal{M}}\left\{\sum_{n\in\mathcal{N}}a_{n,m}D_n(t)\left(\frac{\sum_{l=1}^{l_n(t)}\chi_n^l}{\phi_m^\text{G} f_m^\text{G}}+\right.\right.\nonumber\\&\left.\left.\quad\quad\frac{o_n^{l_n(t)}}{B\log\left(1+\frac{P_mH_m(t)}{\eta_m(t)+N_0B}\right)}+\frac{\sum_{l=l_n(t)+1}^{L_n}\chi_n^l}{\phi_m^\text{A} f_m^\text{A}(t)}\right)\right\}\\
	\text{s.t.}\quad 
	&\textbf{C2}, \textbf{C5}.\nonumber
\end{alignat}Similarly, the sub-problem in (\ref{P2_2}) can be solved by the bisection method. Let $g_2^*(t)$ denote the minimum value of the objective function in (\ref{P2_2}). From the constraint \textbf{C2}, the lower bound of $g_2^*(t)$ can be derived as\begin{small}\begin{alignat}{1}
g_2^\text{min}(t)&=\max_{m\in\mathcal{M}}\left\{\sum_{n\in\mathcal{N}}a_{n,m}D_n(t)\left(\frac{\sum_{l=1}^{l_n(t)}\chi_n^l}{\phi_m^\text{G} f_m^\text{G}}+\frac{\sum_{l=l_n(t)+1}^{L_n}\chi_n^l}{\phi_m^\text{A} \sum\limits_{j\in\mathcal{J}}b_{m,j}f_j^\text{\rm{max}}}\right.\right.\nonumber\\&\left.\left.+\frac{o_n^{l_n(t)}}{B\log\left(1+\frac{P_mH_m(t)}{\eta_m(t)+N_0B}\right)}\right)\right\}.
\end{alignat}\end{small}Let $\boldsymbol{\epsilon}=\{\epsilon_m\}_{m=1}^M$ denote the set of any possible $f_m^\text{\rm{A}}(t)$ that satisfies the constraints \textbf{C2} and \textbf{C5}, i.e., $f_j^\text{\rm{min}}\leq\sum_{m\in\mathcal{M}}b_{m,j}$ $\epsilon_m\leq f_j^\text{\rm{max}}$, and $0\leq\sum_{m\in\mathcal{M}}b_{m,j}\frac{v_m^\text{A}\left(\epsilon_m\right)^2}{\phi_m^\text{A}}\sum_{n\in\mathcal{N}}a_{n,m}D_n(t)$ $\sum_{l=l_n(t)+1}^{L_n}\chi_n^l+e_j^\text{bloc}(t)\leq E_j^\text{\rm{A}}(t)$. The upper bound of $g_2^*(t)$ can be derived as\begin{small}\begin{alignat}{1}
g_2^\text{max}(t)&=\max_{m\in\mathcal{M}}\left\{\sum_{n\in\mathcal{N}}a_{n,m}D_n(t)\left(\frac{\sum_{l=1}^{l_n(t)}\chi_n^l}{\phi_m^\text{G} f_m^\text{G}}+\frac{\sum_{l=l_n(t)+1}^{L_n}\chi_n^l}{\phi_m^\text{A} \epsilon_m}\right.\right.\nonumber\\&\left.\left.+\frac{o_n^{l_n(t)}}{B\log\left(1+\frac{P_mH_m(t)}{\eta_m(t)+N_0B}\right)}\right)\right\}.
\end{alignat}\end{small}In each iteration, we first calculate the mid point of the interval between the lower and upper bound of $g_2^*(t)$, i.e., $\lambda=\frac{1}{2}\left(g_2^\text{min}(t)+g_2^\text{max}(t)\right)$, and then compute the upper bound of $f_m^\text{A}(t)$ according to the equation as\begin{small}\begin{alignat}{1}
\lambda&=\max_{m\in\mathcal{M}}\left\{\sum\limits_{n\in\mathcal{N}}a_{n,m}D_n(t)\left(\frac{\sum_{l=1}^{l_n(t)}\chi_n^l}{\phi_m^\text{G} f_m^\text{G}}+\frac{o_n^{l_n(t)}}{B\log\left(1+\frac{P_mH_m(t)}{\eta_m(t)+N_0B}\right)}\right.\right.\nonumber\\&\left.\left.+\frac{\sum_{l=l_n(t)+1}^{L_n}\chi_n^l}{\phi_m^\text{A} f_m^\text{A}(t)}\right)\right\}.\end{alignat}\end{small}The upper bound of $f_m^\text{A}(t)$ is derived as\begin{small}\begin{alignat}{1}\label{defvg}
f_m^\text{A}(t)&\ge\frac{\sum_{l=l_n(t)+1}^{L_n}\chi_n^l}{\phi_m^\text{A}}\left(\lambda-\sum\limits_{n\in\mathcal{N}}a_{n,m}D_n(t)\left(\frac{\sum_{l=1}^{l_n(t)}\chi_n^l}{\phi_m^\text{G} f_m^\text{G}}\right.\right.\nonumber\\&\left.\left.+\frac{o_n^{l_n(t)}}{B\log\left(1+\frac{P_mH_m(t)}{\eta_m(t)+N_0B}\right)}\right)\right)^{-1}.\end{alignat}\end{small}If the derived upper bound of $f_m^\text{A}(t)$ in (\ref{defvg}) satisfies the constraints \textbf{C2} and \textbf{C5}, the interval between the lower and upper bound of $g_2^*(t)$ can be narrowed down by updating $g_2^\text{max}(t)=\lambda$. Otherwise, the lower bound of $g_2^*(t)$ is updated by $g_2^\text{min}(t)=\lambda$. Thus, the optimal computation frequency for DNN reference is derived as\begin{small}\begin{alignat}{1}
f_m^\text{A}(t)&=\frac{\sum_{l=l_n(t)+1}^{L_n}\chi_n^l}{\phi_m^\text{A}}\left(\lambda-\sum\limits_{n\in\mathcal{N}}a_{n,m}D_n(t)\left(\frac{\sum_{l=1}^{l_n(t)}\chi_n^l}{\phi_m^\text{G} f_m^\text{G}}\right.\right.\nonumber\\&\left.\left.+\frac{o_n^{l_n(t)}}{B\log\left(1+\frac{P_mH_m(t)}{\eta_m(t)+N_0B}\right)}\right)\right)^{-1}.\end{alignat}\end{small}

Fixing the computation frequency $\boldsymbol{f}^\text{A}(t)$ for the top-layer DNN inference and the computation frequency $\boldsymbol{f}^\text{bloc}(t)$ for block generation, the problem in \textbf{P2} can be rewritten as\begin{alignat}{1}\label{P2_3}
	\min_{\boldsymbol{l}(t)}\;&g_3\left(\boldsymbol{l}(t)\right)=\sum_{j\in\mathcal{J}}\left(S_j(t)-Q_j(t)\right)U_j(t)+V\left(\max_{m\in\mathcal{M}}\left\{\sum_{n\in\mathcal{N}}\right.\right.\nonumber\\&a_{n,m}D_n(t)\left(\frac{\sum_{l=1}^{l_n(t)}\chi_n^l}{\phi_m^\text{G} f_m^\text{G}}+\frac{o_n^{l_n(t)}}{B\log\left(1+\frac{P_mH_m(t)}{\eta_m(t)+N_0B}\right)}\right.\nonumber\\&\left.\left.\left.+\frac{\sum_{l=l_n(t)+1}^{L_n}\chi_n^l}{\phi_m^\text{A} f_m^\text{A}(t)}\right)\right\}+\frac{-\ln(1-p_0)}{\sum_{j\in\mathcal{J}}f_j^\text{bloc}(t)e^{\alpha U_j(t)+\beta}}\right)\!\\
	\text{s.t.}\quad 
	&\textbf{C1}, \textbf{C4}, \textbf{C5}.\nonumber
\end{alignat}We employ the approach of case analysis to solve the non-linear integer programming problem in (\ref{P2_3}). That is, we first split the problem in (\ref{P2_3}) into $M$ sub-problems, and then solve each sub-problem separately. Notably, the optimal solution of the original problem in (\ref{P2_3}) belongs to the union of the solutions to the sub-problems. Thus, the optimal DNN partitioning point $l_n(t)$ can be obtained by comparing the solutions to the sub-problems. Let $\mathcal{I}=\{1,2,...,M\}$ denote the index set of the $M$ sub-problems. In the $i$-th sub-problem, we assume that the $i$-th gateway and its associated AP take the largest amount of time for DNN inference, i.e., $\sum_{n\in\mathcal{N}}a_{n,m}D_n(t)\left(\frac{\sum_{l=1}^{l_m(t)}\chi_m^l}{\phi_m^\text{G} f_m^\text{G}}+\frac{o_m^{l_m(t)}}{B\log\left(1+\frac{P_mH_m(t)}{\eta_m(t)+N_0B}\right)}+\frac{\sum_{l=l_m(t)+1}^{L_m}\chi_m^l}{\phi_m^\text{A} f_m^\text{A}(t)}\right)$ $\leq \sum_{n\in\mathcal{N}}a_{n,i}D_n(t)\left(\frac{\sum_{l=1}^{l_i(t)}\chi_i^l}{\phi_i^\text{G} f_i^\text{G}}+\frac{o_i^{l_i(t)}}{B\log\left(1+\frac{P_iH_i(t)}{\eta_i(t)+N_0B}\right)}+\frac{\sum_{l=l_i(t)+1}^{L_i}\chi_i^l}{\phi_i^\text{A} f_i^\text{A}(t)}\right)$, $\forall m\in \mathcal{M}$. Thus, the $i$-th sub-problem can be written as\begin{alignat}{1}\label{P2_4}
&\!\!\!\!\!\!\!\!\!\!\!\!\!\!\min_{\boldsymbol{l}(t)}\;g_3^i\left(\boldsymbol{l}(t)\right)=V\!\!\sum_{n\in\mathcal{N}}\!\frac{a_{n,i}D_n(t)o_n^{l_n(t)}}{B\log\left(1\!+\!\frac{P_iH_i(t)}{\eta_i(t)+N_0B}\right)}+\!\!\sum_{n\in\mathcal{N}}\!\Bigg(Va_{n,i}\!\left(\frac{1}{\phi_i^\text{G} f_i^\text{G}}\right.\nonumber\\&\!\!\!\!\!\!\!\!\!\!\!\!\!\!\left.-\frac{1}{\phi_i^\text{A} f_i^\text{A}(t)}\right)-\!\!\sum_{j\in\mathcal{J}}\!\sum_{m\in\mathcal{M}}\!\!\left(S_j(t)-Q_j(t)\right)b_{m,j} a_{n,m}\Bigg)D_n(t)\!\sum_{l=1}^{l_n(t)}\!\chi_n^l\nonumber\\&\!\!\!\!\!\!\!\!\!\!\!\!\!\!+\frac{-V\ln\left(1-p_0\right)}{\sum_{j\in\mathcal{J}}e^\beta f_j^\text{bloc}(t) \frac{e^{\alpha\sum_{n\in\mathcal{N}}\sum_{m\in\mathcal{M}}b_{m,j}a_{n,m}D_n(t)\sum_{l=1}^{L_n}\chi_n^l}}{e^{\alpha\sum_{n\in\mathcal{N}}\sum_{m\in\mathcal{M}}b_{m,j}a_{n,m}D_n(t)\sum_{l=1}^{l_n(t)}\chi_n^l}}}+\!\!\sum_{j\in\mathcal{J}}\!\!\left(S_j(t)\right.\nonumber\\&\!\!\!\!\!\!\!\!\!\!\!\!\!\!\left.- \,Q_j(t)\right)\left(\sum_{m\in\mathcal{M}}\sum_{n\in\mathcal{N}}b_{m,j}a_{n,m}D_n(t)\sum_{l=1}^{L_n}\chi_n^l\right)+V\sum_{n\in\mathcal{N}}\frac{1}{\phi_i^\text{A} f_i^\text{A}(t)}\nonumber\\&\!\!\!\!\!\!\!\!\!\!\!\!\!\!\times a_{n,i}D_n(t)\sum_{l=1}^{L_n}\chi_n^l\\
\text{s.t.}\quad 
&\textbf{C1}, \textbf{C4}, \textbf{C5},\nonumber\\
&\textbf{C9}: \sum_{n\in\mathcal{N}}a_{n,m}D_n(t)\left(\frac{\sum_{l=1}^{L_n}\chi_n^l}{\phi_m^\text{A} f_m^\text{A}(t)}+\left(\frac{1}{\phi_m^\text{G} f_m^\text{G}}-\frac{1}{\phi_m^\text{A} f_m^\text{A}(t)}\right)\right.\nonumber\\&\left.\sum_{l=1}^{l_n(t)}\chi_n^l+\frac{o_n^{l_n(t)}}{B\log\left(1+\frac{P_mH_m(t)}{\eta_m(t)+N_0B}\right)}\right)\leq \sum_{n\in\mathcal{N}}a_{n,i}D_n(t)\nonumber\\&\left(\left(\frac{1}{\phi_i^\text{G} f_i^\text{G}}-\frac{1}{\phi_i^\text{A} f_i^\text{A}(t)}\right)\sum_{l=1}^{l_n(t)}\chi_n^l+\frac{o_n^{l_n(t)}}{B\log\left(1+\frac{P_iH_i(t)}{\eta_i(t)+N_0B}\right)}\right.\nonumber\\&\left.+\frac{\sum_{l=1}^{L_n}\chi_n^l}{\phi_i^\text{A} f_i^\text{A}(t)}\right), \forall m\in \mathcal{M}, t\in \mathcal{T}.\nonumber
\end{alignat}We solve the non-convex nonlinear integer optimization problem in (\ref{P2_4}) by employing the branch and bound method. Then, given the optimized DNN partitioning point $\boldsymbol{l}^i(t)$ in each sub-problem, we have $i^*=\arg\min_{i\in\mathcal{I}}\{g_3^i\left(\boldsymbol{l}^i(t)\right)\}$. Therefore, the optimal DNN partitioning point is given by $\boldsymbol{l}^*(t)=\boldsymbol{l}^{i^*}(t)$.

\section{Performance Analysis}\label{Performance Analysis}
In this section, we analyze the performance of the proposed DPRA algorithm in terms of asymptotic optimality, and characterize the trade-off between minimizing system latency and satisfying long-term off-chain reputation constraint.
\begin{theorem}\label{performance analysis}
	Let $\psi^\text{opt}$ and $\psi^*$ denote the system latency under the optimal solution of \textbf{P0} and \textbf{P2}, we have
	\begin{alignat}{1}\label{equ2:theorem1}
		\psi^*-\psi^\text{opt}\leq \frac{H}{V}+\frac{\mathbb{E}\{L(0)-L(T)\}}{VT},\end{alignat}\begin{alignat}{1}\label{equ3:theorem1}
		\frac{1}{T}\sum_{t=0}^{T-1}U_j(t)&\ge U^\text{min}-\left( \frac{H+V\left(\psi^\text{opt}-\tau^\text{min}\right)}{T}\right.\nonumber\\&\left.+\sum_{j\in \mathcal{J}}\frac{\mathbb{E}\left\{\left(Q_j(0)\right)^2+\left(S_j(0)\right)^2\right\}}{T^2}\right)^{\frac{1}{2}},
	\end{alignat}and\begin{alignat}{1}\label{equ5:theorem1}
	\frac{1}{T}\sum_{t=0}^{T-1}U_j(t)&\leq U^\text{max}+\left( \frac{H+V\left(\psi^\text{opt}-\tau^\text{min}\right)}{T}\right.\nonumber\\&\left.+\sum_{j\in \mathcal{J}}\frac{\mathbb{E}\left\{\left(Q_j(0)\right)^2+\left(S_j(0)\right)^2\right\}}{T^2}\right)^{\frac{1}{2}},
\end{alignat}where\begin{small}\begin{alignat}{1}&\tau^\text{min}=\max_{m\in\mathcal{M}}\left\{\frac{\sum_{n\in\mathcal{N}}a_{n,m}\overline{D}_n\sum_{l=1}^{L_n}\chi_n^{l}}{\max\Big\{\max\limits_{m\in\mathcal{M}}\left\{\phi_m^\text{G} f_m^\text{G}\right\},\max\limits_{m\in\mathcal{M}}\left\{\phi_m^\text{A}\right\}\max\limits_{j\in\mathcal{J}}\left\{f_j^\text{max}\right\}\Big\}}\right.\nonumber\\&\left.+\frac{\sum\limits_{n\in\mathcal{N}}a_{n,m}\overline{D}_n\min_{l\in\mathcal{L}_n}\left\{o_n^l\right\}}{B\log\left(1+\frac{P_m\overline{H_m}}{\overline{\eta_m}+N_0B}\right)}\right\}+\frac{-\ln(1-p_0)}{\max_{j\in\mathcal{J}}\{f_j^\text{max}\}}\bigg({\sum}_{j\in\mathcal{J}}e^{\beta}\nonumber\\&\times e^{\alpha g\left(\sum_{m\in\mathcal{M}}\sum_{n\in\mathcal{N}}b_{m,j}a_{n,m}\overline{D}_n\sum_{l=1}^{L_n}\chi_n^l\right)}\bigg)^{-1}.
\end{alignat}\end{small}
\end{theorem}
\begin{IEEEproof}
	Please see Appendix B.
\end{IEEEproof}

\textbf{Theorem} \ref{performance analysis} presents an [$\mathcal{O}(1/V)$, $\mathcal{O}(V)$] trade-off between minimizing the system latency and satisfying the long-term off-chain reputation constraint \textbf{C6}. From (\ref{equ2:theorem1}), it can be observed that the DPRA algorithm converges to the optimal solution as the Lyapunov control parameter $V$ increases, which verifies the asymptotic optimality of the proposed algorithm. From (\ref{equ3:theorem1}) and (\ref{equ5:theorem1}), the average off-chain reputation approaches the lower and upper bound specified in constraint \textbf{C6} as $V$ decreases. That is, a large value of $V$ encourages reducing the system latency, which can be employed for real-time delay-sensitive IIoT applications. Conversely, a small value of $V$ pushes the average off-chain reputation to satisfy the long-term off-chain reputation constraint, thereby enhancing the trustworthiness of the B-DT system.
\section{Experimental Results}\label{Experimental Results}
In this section, we evaluate the performance of the proposed DPRA algorithm. Consider that the B-DT system consists of $N=60$ devices, $M=20$ gateways, and $J=4$ APs. Each gateway is associated with $3$ devices, and each AP is associated with $5$ gateways. The APs are equally divided into two types. For \textbf{Type 1} APs and the associated devices, we set the energy arrivals $E_j^\text{A,max}=10$ J and the data arrivals $\Theta_n=100$. For \textbf{Type 2} APs and the associated devices, we set $E_j^\text{A,max}=30$ J and $\Theta_n=50$. For each AP, $v_j^\text{A}=10^{-24}$, $\phi_m^\text{A}=32$ FLOPs per CPU cycle, and $f_j^\text{max}=0.1$ GHz. For each gateway, $E_j^\text{A,max}=0.5$ J, $f_m^\text{G}$ is uniformly distributed within $[1,10]$ MHz, $\phi_m^\text{G}=8$ FLOPs per CPU cycle, $v_m^G=v_j^\text{A}=10^{-24}$, $P_m=100$ mW, and $d_m$ is uniformly distributed within $[0,50]$ m. The channel parameters are set as $d_0=1$ m, $h_0=10^{-3}$, $B=5$ MHz, and $N_0=-174$ dBm/Hz. The uplink inference $\eta_m(t)$ is produced by the Gaussian distribution, and the channel power gain $\rho_m(t)$ is exponentially distributed with unit mean. Besides, we set $p_0=1-10^{-15}$, $\alpha=5\times 10^{-5}$, $\beta=-29$, $U^\text{min}=25$, and $U^\text{min}=75$. In addition, we adopt VGG-11 on Cifar-10 dataset and CNN on Fashion-MNIST dataset to demonstrate the DTs' DNN inference tasks. 

\begin{figure}[!t]
	\centering
	\includegraphics[width=3.2in]{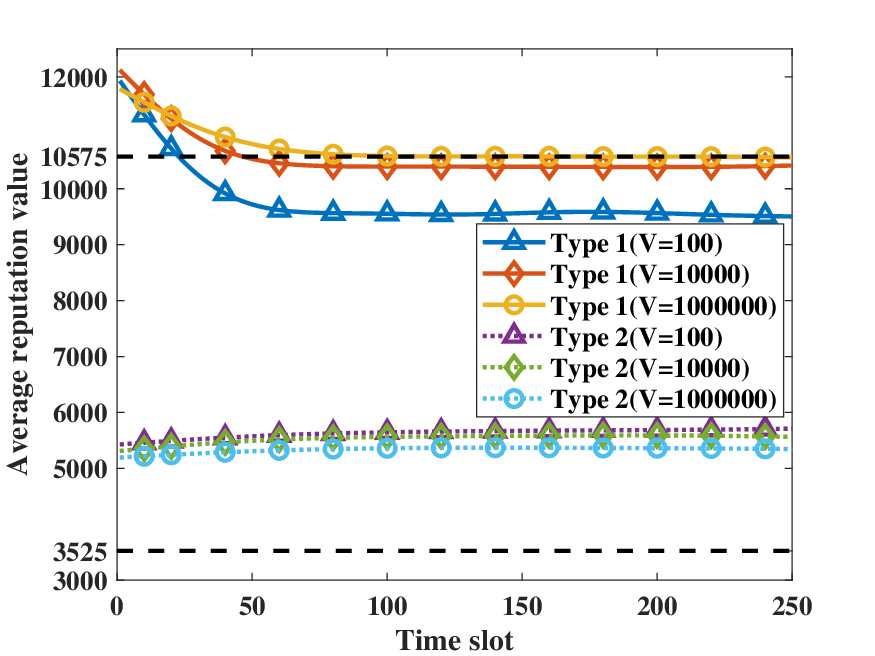}
	\caption{The average reputation value of the APs under the DPRA algorithm versus time slot.}
	\label{fig6}
\end{figure}

Fig. \ref{fig6} plots the time variation of the average off-chain reputation value for \textbf{Type 1} and \textbf{Type 2} APs under the DPRA algorithm with the Lyapunov control parameter $V=10^2$, $10^4$, and $10^6$. First, we observe that the average off-chain reputation values of \textbf{Type 1} and \textbf{Type 2} APs fall within the lower and upper limits of the long-term off-chain reputation value specified in constraint \textbf{C6} when $t\ge100$. Second, it can be seen that the average off-chain reputation value for \textbf{Type 1} APs decreases in the beginning and finally falls below the upper limit of the long-term off-chain reputation value $U^\text{max}$. In addition, the gap between the average off-chain reputation value of \textbf{Type 2} APs and the lower limit of the long-term off-chain reputation value $U^\text{min}$ gradually increases over time. This indicates that the average off-chain reputation values of \textbf{Type 1} and \textbf{Type 2} APs gradually approach the mid point of the lower and upper limits of the off-chain reputation value as the time elapses, which is consistent with (\ref{equ3:theorem1}) and (\ref{equ5:theorem1}) in \textbf{Theorem} \ref{performance analysis}. 

\begin{figure}[!t]
	\centering
	\includegraphics[width=3.2in]{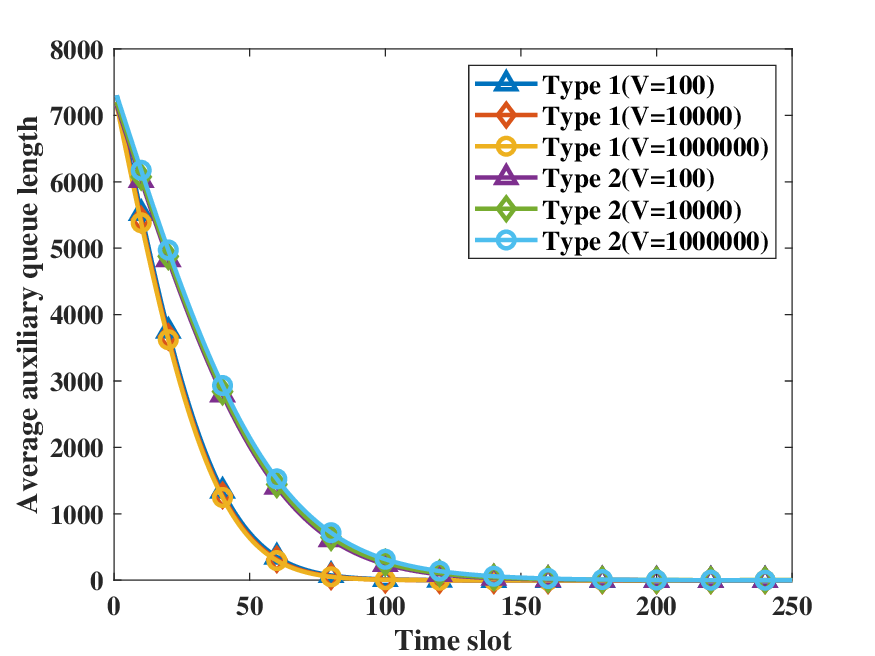}
	\caption{The average auxiliary queue $\boldsymbol{Q}(t)$ under the DPRA algorithm versus time slot.}
	\label{fig5}
\end{figure}
\begin{figure}[!t]
	\centering
	\includegraphics[width=3.2in]{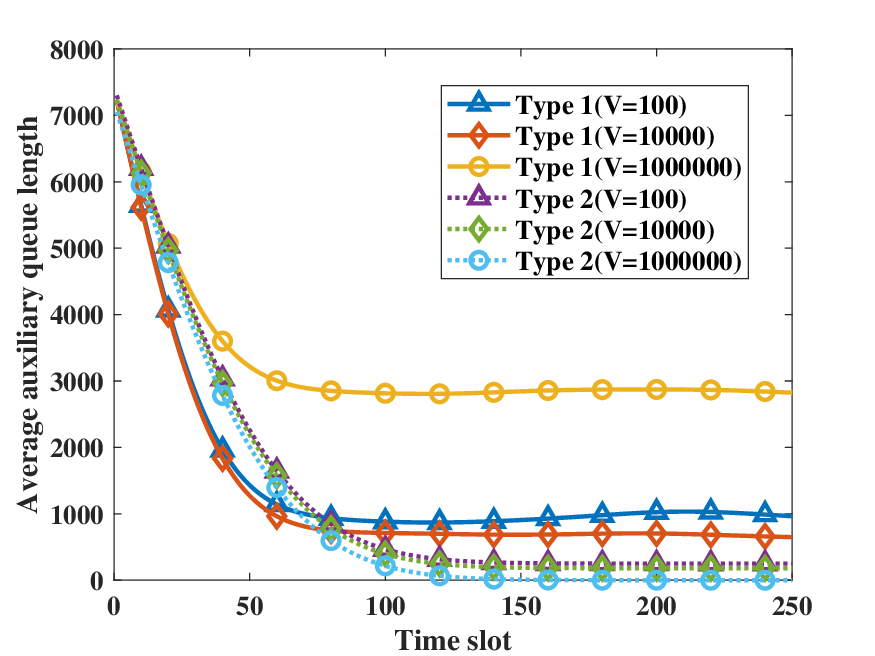}
	\caption{The average auxiliary queue $\boldsymbol{S}(t)$ under the DPRA algorithm versus time slot.}
	\label{fig4}
\end{figure}

Figs. \ref{fig5} and \ref{fig4} illustrate the time variation of the average auxiliary queue lengths for \textbf{Type 1} and \textbf{Type 2} APs under the DPRA algorithm with the Lyapunov control parameter $V=10^2$, $10^4$, and $10^6$, respectively. First, it can be observed that the average auxiliary queue lengths of $\boldsymbol{Q}(t)$ and $\boldsymbol{S}(t)$ decrease in the beginning and then stabilize as the time elapses. Second, the average auxiliary queue lengths $\boldsymbol{Q}(t)$ and $\boldsymbol{S}(t)$ of \textbf{Type 1} APs decrease faster than \textbf{Type 2} APs. Recall that $Q_j(t+1)=\max\left\{Q_j(t)-U_j(t)+U^\text{min},0\right\}$, and $	S_j(t+1)=\max\left\{S_j(t)+U_j(t)-U^\text{max},0\right\}$ in (\ref{q1}) and (\ref{q2}). This is due to the fact that the average reputation value of \textbf{Type 1} APs is higher than \textbf{Type 2} APs, which is consistent with Fig. \ref{fig6}.

\begin{figure}[!t]
	\centering
	\includegraphics[width=3.2in]{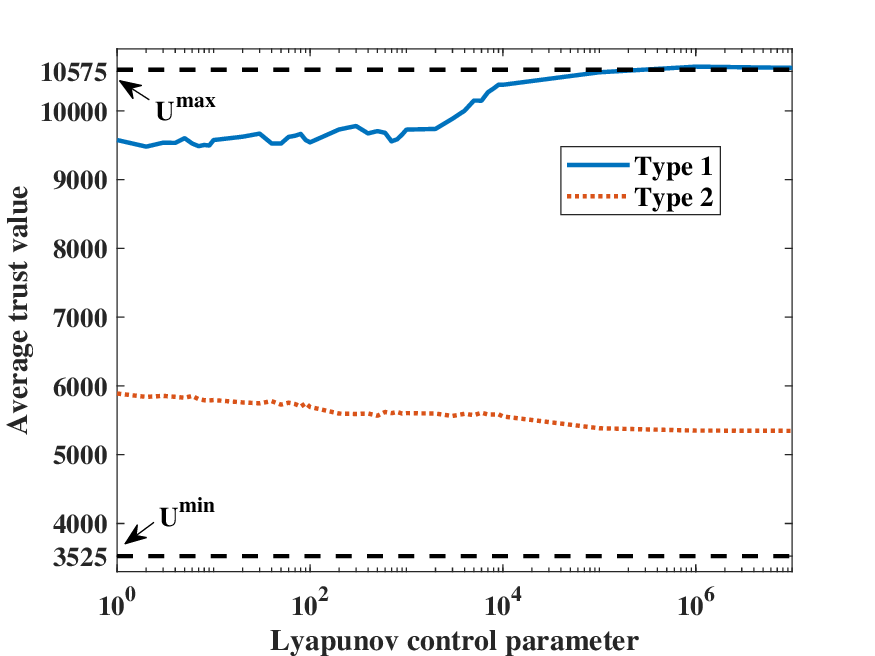}
	\caption{The average reputation values of the APs under the DPRA algorithm versus $V$.}
	\label{fig3}
\end{figure}

Fig. \ref{fig3} plots the average reputation values of \textbf{Type 1} and \textbf{Type 2} APs under the DPRA algorithm over $V\in (0,10^7]$. First, as the control parameter $V$ increase, the average reputation value of \textbf{Type 1} APs approaches to the upper limit of the long-term off-chain reputation value $U^\text{max}$, and the average reputation value of \textbf{Type 2} APs approaches to the lower limit of the long-term off-chain reputation value $U^\text{min}$, respectively. This is due to the fact that $V$ tunes the trade-off between minimizing the system latency and satisfying the long-term off-chain reputation constraint \textbf{C6}. A small value of $V$ promotes the minimization of system latency and ignores the satisfaction of the long-term off-chain reputation constraint. Second, the average reputation value of \textbf{Type 1} APs is higher than that of \textbf{Type 2} APs. That is, the associated gateways offload more DNN inference tasks to \textbf{Type 1} APs than \textbf{Type 2} APs. This is because that the rate of new data arrivals of the DTs maintained at the \textbf{Type 1} APs' associated gateways is higher than that of the \textbf{Type 2} APs.

\begin{figure}[!t]
	\centering
	\includegraphics[width=3.2in]{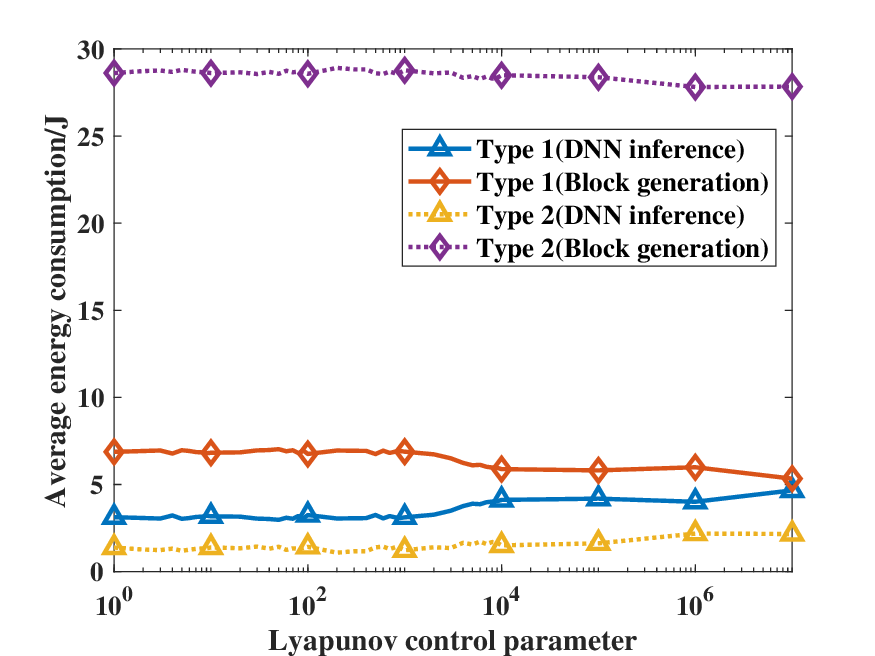}
	\caption{The average energy consumption comparison between DNN inference and block generation under DPRA.}
	\label{fig2}
\end{figure}

Fig. \ref{fig2} plots the average energy consumption of \textbf{Type 1} and \textbf{Type 2} APs for DNN inference and block generation under the DPRA algorithm over the Lyapunov control parameter $V\in (0,10^7]$. First, the energy consumption of \textbf{Type 1} APs for DNN inference tasks are higher than that of \textbf{Type 2} APs. It is because that the associated gateways offload more DNN inference tasks to \textbf{Type 1} APs than \textbf{Type 2} APs, which is consistent with Fig. \ref{fig6}. Second, we notice that the energy consumption of  \textbf{Type 1} and \textbf{Type 2} APs for DNN inference tasks increases with $V$, while the energy consumption for block generation decreases with $V$. This is due to the fact that a large value of $V$ can lead to a large reputation value for the APs, which contributes to a low block generation difficulty and consequently reduces the energy consumption for block generation.

\begin{figure}[!t]
	\centering
	\includegraphics[width=3.1in]{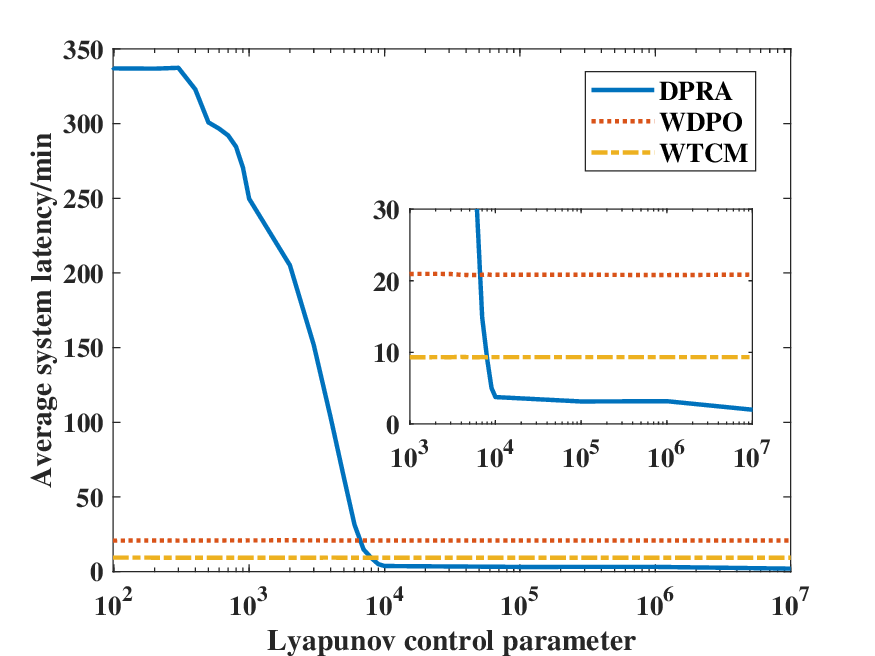}
	\caption{The average system latency comparison between DPRA and baselines.}
	\label{fig1}
\end{figure}

Fig. \ref{fig1} shows the average system latency under the DPRA algorithm over the Lyapunov control parameter $V\in (0,10^7]$. For comparison purpose, we also simulate two baselines as follows: (a) computation resource allocation policy without DNN partitioning point optimization (WDPO), and (b) DNN partitioning point optimization and computation resource allocation policy without reputation based consensus mechanism (WTCM). First, it can be found that the average system latency of the proposed DPRA decreases as $V$ increase. It reveals that a larger value of $V$ can lead to a smaller system latency, which conforms to \textbf{Theorem} \ref{performance analysis}. Second, DPRA shows a lower system latency than baseline schemes. Compared with WDPS, DPRA reduces system latency by jointly optimizing the DNN partitioning point and the computation frequency in each time slot. Compared with WTCM, DPRA adjusts the block generation difficulty according to the off-chain reputation, which reduces the system latency while guaranteeing the trustworthiness of the B-DT system.

\section{Conclusion}\label{Conclusion}
In this paper, we have developed a communication and computation efficient B-DT framework for wireless IIoT networks with DNN partitioning method. By adaptively tuning the block generation difficulty according to APs' computation resource contributions to the DNN inference tasks, the proposed consensus mechanism can reduce the block generation latency while ensuring the system trustworthiness. By jointly optimizing the DNN partitioning point and computation resource allocation for DNN inference and block generation, the proposed DPRA algorithm can minimize the system latency with the time-varying channel state and energy arrivals. The long-term off-chain reputation constraint guarantees both scalability and security of the B-DT system. The [$\mathcal{O}(1/V)$, $\mathcal{O}(V)$] trade-off between the minimization of system latency and the satisfaction of the long-term off-chain reputation constraint indicates that minimizing the system latency and improving the system trustworthiness can be balanced by adjusting the Lyapunov control parameter $V$. Experimental results show that the DPRA algorithm outperforms the baselines in terms of reducing the overall latency while guaranteeing the trustworthiness of the B-DT system.
\bibliographystyle{IEEEtran}
\bibliography{references}

\begin{thebibliography}{10}
\providecommand{\url}[1]{#1}
\csname url@samestyle\endcsname
\providecommand{\newblock}{\relax}
\providecommand{\bibinfo}[2]{#2}
\providecommand{\BIBentrySTDinterwordspacing}{\spaceskip=0pt\relax}
\providecommand{\BIBentryALTinterwordstretchfactor}{4}
\providecommand{\BIBentryALTinterwordspacing}{\spaceskip=\fontdimen2\font plus
\BIBentryALTinterwordstretchfactor\fontdimen3\font minus \fontdimen4\font\relax}
\providecommand{\BIBforeignlanguage}[2]{{%
\expandafter\ifx\csname l@#1\endcsname\relax
\typeout{** WARNING: IEEEtran.bst: No hyphenation pattern has been}%
\typeout{** loaded for the language `#1'. Using the pattern for}%
\typeout{** the default language instead.}%
\else
\language=\csname l@#1\endcsname
\fi
#2}}
\providecommand{\BIBdecl}{\relax}
\BIBdecl

\bibitem{DBLP:journals/iotj/WuZ021}
Y.~Wu, K.~Zhang, and Y.~Zhang, ``Digital twin networks: {A} survey,'' \emph{{IEEE} Internet Things J.}, vol.~8, no.~18, pp. 13\,789--13\,804, 2021.

\bibitem{DBLP:journals/tcyb/GuoZRH23}
D.~Guo, R.~Y. Zhong, Y.~Rong, and G.~G.~Q. Huang, ``Synchronization of shop-floor logistics and manufacturing under {IIoT} and digital twin-enabled graduation intelligent manufacturing system,'' \emph{{IEEE} Trans. Cybern.}, vol.~53, no.~3, pp. 2005--2016, 2023.

\bibitem{DBLP:conf/globecom/HuLGZS22}
S.~Hu, M.~Li, J.~Gao, C.~Zhou, and X.~S. Shen, ``Digital twin-assisted adaptive {DNN} inference in industrial internet of things,'' in \emph{{IEEE} {GLOBECOM}, Rio de Janeiro, Brazil, December 4-8, 2022}, pp. 1025--1030.

\bibitem{DBLP:journals/tii/LuHZMZ21a}
Y.~Lu, X.~Huang, K.~Zhang, S.~Maharjan, and Y.~Zhang, ``Communication-efficient federated learning for digital twin edge networks in industrial {IoT},'' \emph{{IEEE} Trans. Ind. Informatics}, vol.~17, no.~8, pp. 5709--5718, 2021.

\bibitem{DBLP:journals/tii/GuoTK23}
Q.~Guo, F.~Tang, and N.~Kato, ``Federated reinforcement learning-based resource allocation for {D2D}-aided digital twin edge networks in {6G} industrial {IoT},'' \emph{{IEEE} Trans. Ind. Informatics}, vol.~19, no.~5, pp. 7228--7236, 2023.

\bibitem{DBLP:journals/sensors/ChukhnoCACIM20}
O.~Chukhno, N.~Chukhno, G.~Araniti, C.~Campolo, A.~Iera, and A.~Molinaro, ``Optimal placement of social digital twins in edge {IoT} networks,'' \emph{Sensors}, vol.~20, no.~21, p. 6181, 2020.

\bibitem{DBLP:journals/tvt/SunZWZ20}
W.~Sun, H.~Zhang, R.~Wang, and Y.~Zhang, ``Reducing offloading latency for digital twin edge networks in {6G},'' \emph{{IEEE} Trans. Veh. Technol.}, vol.~69, no.~10, pp. 12\,240--12\,251, 2020.

\bibitem{DBLP:journals/jcin/DaiZ22}
Y.~Dai and Y.~Zhang, ``Adaptive digital twin for vehicular edge computing and networks,'' \emph{J. Commun. Inf. Networks}, vol.~7, no.~1, pp. 48--59, 2022.

\bibitem{DBLP:journals/tii/ZhouJLLMGT22}
Z.~Zhou, Z.~Jia, H.~Liao, W.~Lu, S.~Mumtaz, M.~Guizani, and M.~Tariq, ``Secure and latency-aware digital twin assisted resource scheduling for {5G} edge computing-empowered distribution grids,'' \emph{{IEEE} Trans. Ind. Informatics}, vol.~18, no.~7, pp. 4933--4943, 2022.

\bibitem{DBLP:journals/iotj/WangLSLMZ23}
D.~Wang, B.~Li, B.~Song, Y.~Liu, K.~Muhammad, and X.~Zhou, ``Dual-driven resource management for sustainable computing in the blockchain-supported digital twin {IoT},'' \emph{{IEEE} Internet Things J.}, vol.~10, no. 8, April 15, pp. 6549--6560, 2023.

\bibitem{DBLP:conf/msn/WuXL21}
J.~Wu, Q.~Xia, and Q.~Li, ``Efficient privacy-preserving federated learning for resource-constrained edge devices,'' in \emph{{MSN} 2021, Exeter, United Kingdom, December 13-15, 2021}, pp. 191--198.

\bibitem{DBLP:journals/cn/WangXXJL23}
Z.~Wang, H.~Xu, Y.~Xu, Z.~Jiang, and J.~Liu, ``{CoopFL}: Accelerating federated learning with {DNN} partitioning and offloading in heterogeneous edge computing,'' \emph{Comput. Networks}, vol. 220, p. 109490, 2023.

\bibitem{DBLP:journals/jsac/DengLMWSDC23}
X.~Deng, J.~Li, C.~Ma, K.~Wei, L.~Shi, M.~Ding, and W.~Chen, ``Low-latency federated learning with {DNN} partition in distributed industrial {IoT} networks,'' \emph{{IEEE} J. Sel. Areas Commun.}, vol.~41, no.~3, pp. 755--775, 2023.

\bibitem{DBLP:journals/tmc/GaoSSQLL23}
M.~Gao, R.~Shen, L.~Shi, W.~Qi, J.~Li, and Y.~Li, ``Task partitioning and offloading in {DNN}-task enabled mobile edge computing networks,'' \emph{{IEEE} Trans. Mob. Comput.}, vol.~22, no.~4, pp. 2435--2445, Apr. 2023.

\bibitem{DBLP:journals/jcloudc/GuoZS23}
C.~Guo, L.~Zhengqing, and J.~Song, ``A privacy protection approach in edge-computing based on maximized {DNN} partition strategy with energy saving,'' \emph{J. Cloud Comput.}, vol.~12, no.~1, p.~29, Mar. 2023.

\bibitem{DBLP:journals/jstsp/WangCL23}
F.~Wang, S.~Cai, and V.~K.~N. Lau, ``Decentralized {DNN} task partitioning and offloading control in {MEC} systems with energy harvesting devices,'' \emph{{IEEE} J. Sel. Top. Signal Process.}, vol.~17, no.~1, pp. 173--188, Jan. 2023.

\bibitem{DBLP:journals/ipm/PutzDEP21}
B.~Putz, M.~Dietz, P.~Empl, and G.~Pernul, ``Ethertwin: Blockchain-based secure digital twin information management,'' \emph{Inf. Process. Manag.}, vol.~58, no.~1, p. 102425, 2021.

\bibitem{DBLP:journals/iotj/WangCL23}
C.~Wang, Z.~Cai, and Y.~Li, ``Sustainable blockchain-based digital twin management architecture for {IoT} devices,'' \emph{{IEEE} Internet Things J.}, vol.~10, no. 8, April 15, pp. 6535--6548, 2023.

\bibitem{DBLP:journals/tii/LuHZMZ21}
Y.~Lu, X.~Huang, K.~Zhang, S.~Maharjan, and Y.~Zhang, ``Low-latency federated learning and blockchain for edge association in digital twin empowered {6G} networks,'' \emph{{IEEE} Trans. Ind. Informatics}, vol.~17, no.~7, pp. 5098--5107, 2021.

\bibitem{DBLP:journals/iotj/YangYLZL19}
Z.~Yang, K.~Yang, L.~Lei, K.~Zheng, and V.~C.~M. Leung, ``Blockchain-based decentralized trust management in vehicular networks,'' \emph{{IEEE} Internet Things J.}, vol.~6, no.~2, pp. 1495--1505, 2019.

\bibitem{DBLP:journals/tetc/ZhangLZWK21}
H.~Zhang, J.~Liu, H.~Zhao, P.~Wang, and N.~Kato, ``Blockchain-based trust management for internet of vehicles,'' \emph{{IEEE} Trans. Emerg. Top. Comput.}, vol.~9, no.~3, pp. 1397--1409, 2021.

\bibitem{DBLP:journals/iotj/LuHZMZ21}
Y.~Lu, X.~Huang, K.~Zhang, S.~Maharjan, and Y.~Zhang, ``Communication-efficient federated learning and permissioned blockchain for digital twin edge networks,'' \emph{{IEEE} Internet Things J.}, vol.~8, no.~4, pp. 2276--2288, 2021.

\bibitem{DBLP:journals/network/YaqoobSUJOI20}
I.~Yaqoob, K.~Salah, M.~Uddin, R.~Jayaraman, M.~A. Omar, and M.~Imran, ``Blockchain for digital twins: Recent advances and future research challenges,'' \emph{{IEEE} Netw.}, vol.~34, no.~5, pp. 290--298, 2020.

\bibitem{nakamoto2008bitcoin}
S.~Nakamoto, ``Bitcoin: A peer-to-peer electronic cash system,'' \emph{Decentralized business review}, 2008.

\bibitem{shi2022pooling}
L.~Shi, T.~Wang, J.~Li, S.~Zhang, and S.~Guo, ``Pooling is not favorable: Decentralize mining power of pow blockchain using age-of-work,'' \emph{{IEEE} Trans. on Cloud Comput.}, pp. 1--14, 2022.

\bibitem{DBLP:journals/tpds/LiSWDMSHP22}
J.~Li, Y.~Shao, K.~Wei, M.~Ding, C.~Ma, L.~Shi, Z.~Han, and H.~V. Poor, ``Blockchain assisted decentralized federated learning {(BLADE-FL):} performance analysis and resource allocation,'' \emph{{IEEE} Trans. Parallel Distributed Syst.}, vol.~33, no.~10, pp. 2401--2415, 2022.

\bibitem{DBLP:journals/tsc/FengMMLC23}
X.~Feng, J.~Ma, Y.~Miao, X.~Liu, and K.~R. Choo, ``Regulatable and hardware-based proof of stake to approach nothing at stake and long range attacks,'' \emph{{IEEE} Trans. Serv. Comput.}, vol.~16, no.~3, pp. 2114--2125, 2023.

\bibitem{DBLP:journals/cn/PlattM21}
M.~Platt and P.~McBurney, ``Sybil attacks on identity-augmented proof-of-stake,'' \emph{Comput. Networks}, vol. 199, p. 108424, 2021.

\bibitem{DBLP:journals/ton/ZengCZYZ21}
L.~Zeng, X.~Chen, Z.~Zhou, L.~Yang, and J.~Zhang, ``{CoEdge}: {Cooperative} {DNN} inference with adaptive workload partitioning over heterogeneous edge devices,'' \emph{{IEEE/ACM} Trans. Netw.}, vol.~29, no.~2, pp. 595--608, 2021.

\bibitem{DBLP:journals/tpds/YangBYTZ22}
Z.~Yang, W.~Bao, D.~Yuan, N.~H. Tran, and A.~Y. Zomaya, ``Federated learning with nesterov accelerated gradient,'' \emph{{IEEE} Trans. Parallel Distributed Syst.}, vol.~33, no.~12, pp. 4863--4873, 2022.

\bibitem{DBLP:journals/tcom/PokhrelC20}
S.~R. Pokhrel and J.~Choi, ``Federated learning with blockchain for autonomous vehicles: Analysis and design challenges,'' \emph{{IEEE} Trans. Commun.}, vol.~68, no.~8, pp. 4734--4746, 2020.

\bibitem{DBLP:journals/iotj/QuGLXYLZ20}
Y.~Qu, L.~Gao, T.~H. Luan, Y.~Xiang, S.~Yu, B.~Li, and G.~Zheng, ``Decentralized privacy using blockchain-enabled federated learning in fog computing,'' \emph{{IEEE} Internet Things J.}, vol.~7, no.~6, pp. 5171--5183, 2020.

\bibitem{prokhorov1983probability}
J.~V. Prokhorov and K.~It{\^o}, \emph{Probability theory and mathematical statistics}.\hskip 1em plus 0.5em minus 0.4em\relax Springer, 1983.

\bibitem{DBLP:journals/twc/DengLMWSDCP23}
X.~Deng, J.~Li, C.~Ma, K.~Wei, L.~Shi, M.~Ding, W.~Chen, and H.~V. Poor, ``Blockchain assisted federated learning over wireless channels: Dynamic resource allocation and client scheduling,'' \emph{{IEEE} Trans. Wirel. Commun.}, vol.~22, no.~5, pp. 3537--3553, 2023.

\bibitem{DBLP:journals/tmc/DengLSWZY22}
X.~Deng, J.~Li, L.~Shi, Z.~Wei, X.~Zhou, and J.~Yuan, ``Wireless powered mobile edge computing: Dynamic resource allocation and throughput maximization,'' \emph{{IEEE} Trans. Mob. Comput.}, vol.~21, no.~6, pp. 2271--2288, 2022.

\bibitem{DBLP:series/synthesis/2010Neely}
M.~J. Neely, \emph{Stochastic Network Optimization with Application to Communication and Queueing Systems}, ser. Synthesis Lectures on Communication Networks.\hskip 1em plus 0.5em minus 0.4em\relax Morgan {\&} Claypool Publishers, 2010.

\end{thebibliography}

\clearpage
\appendices
\section{Proof of Lemma \ref{lemma1}}
First, from the auxiliary queue update function in (\ref{s1}) and (\ref{s2}), we have\begin{alignat}{1}\label{R1}
		Q_j(t+1)^2&\leq Q_j(t)^2+\left(U^\text{min}-U_j(t)\right)^2\nonumber\\&+2Q_j(t)\left(U^\text{min}-U_j(t)\right),
\end{alignat}and\begin{alignat}{1}\label{rr2}
S_j(t+1)^2&\leq S_j(t)^2+\Big(U_j(t)-U^\text{max}\Big)^2\nonumber\\&+2S_j(t)\Big(U_j(t)-U^\text{max}\Big).
\end{alignat}Obviously, it can be derived that\begin{alignat}{1}\label{R11}
Q_j(t+1)^2+S_j(t+1)^2&\leq Q_j(t)^2+S_j(t)^2+\Big(U^\text{min}-U_j(t)\Big)^2\nonumber\\&+\Big(U_j^t-U^\text{max}\Big)^2+2Q_j(t)\Big(U^\text{min}-U_j^t\Big)\nonumber\\&+2S_j(t)\Big(U_j(t)-U^\text{max}\Big).
\end{alignat}Next, by moving $Q_j(t)^2$ and $S_j(t)^2$ to the left-hand side of (\ref{R11}), dividing both sides by $2$, summing up the inequalities from $j=1$ to $J$, and taking the conditional expectation, it can be derived that \begin{alignat}{1}\label{R2}&\Delta L(t)\leq \sum_{j\in \mathcal{J}}\mathbb{E}\Big\{Q_j(t)\left(U^\text{min}-U_j(t)\right)+S_j(t)\Big(U_j(t)-U^\text{max}\Big)\nonumber\\&\Big\vert\boldsymbol{\Phi}(t)\Big\}\!+\!\frac{1}{2}\!\sum_{j\in \mathcal{J}}\mathbb{E}\left\{\left.\left(U^\text{min}\right)^2\!+\!\Big(U^\text{max}\Big)^2\!+\!2\left(U_j(t)\right)^2\right\vert\boldsymbol{\Phi}(t)\right\}.\!\!\!
\end{alignat}From (\ref{off-chain}) and constraint \textbf{C1}, it can be derived that $0\leq U_j(t)\leq g\left(\sum_{m\in\mathcal{M}}\sum_{n\in\mathcal{N}}b_{m,j}a_{n,m}D_n(t)\sum_{l=1}^{L_n}\chi_n^l\right)$. Thus, the upper bound of the conditional Lyapunov drift $\Delta(t)$ can be derived as\begin{alignat}{1}
		\Delta L(t)\leq &\sum_{j\in \mathcal{J}}\mathbb{E}\Big\{Q_j(t)\left(U^\text{min}-U_j(t)\right)+S_j(t)\Big(U_j(t)-U^\text{max}\Big)\nonumber\\&\Big\vert\boldsymbol{\Phi}(t)\Big\}+\sum_{j\in \mathcal{J}} g\left(\sum_{m\in\mathcal{M}}\sum_{n\in\mathcal{N}}b_{m,j}a_{n,m}D_n(t)\sum_{l=1}^{L_n}\chi_n^l\right)^2\nonumber\\&+\frac{J}{2}\left(\left(U^\text{min}\right)^2+\Big(U^\text{max}\Big)^2\right),
\end{alignat}which concludes the proof of \textbf{Lemma} \ref{lemma1}.

\section{Proof of Theorem \ref{performance analysis}}
Before we represent the main proof of \textbf{Theorem} \ref{performance analysis}, we first give \textbf{Lemma} \ref{lemma_appendixc} below.

\begin{lemma}\label{lemma_appendixc}
	For any $\delta>0$, there exists an i.i.d. policy $\Omega$ such that
	\begin{alignat}{1}
		\mathbb{E}\left\{\tau(t)\vert\Omega\right\}\leq \psi^\text{opt}+\delta,\end{alignat}and\begin{alignat}{1}
		U^\text{min}-\delta\leq\mathbb{E}\left\{U_j(t)\vert\Omega\right\}\leq U^\text{max}+\delta.
\end{alignat}
\end{lemma}
\begin{IEEEproof}
	Given any $\delta>0$, we can note that there exists a policy $\omega$ which meets all of the constraints in \textbf{P0} and yields that\begin{alignat}{1}\label{euq1:lemma}\lim_{T\to\infty}\inf \left[\frac{1}{T}\sum_{t=0}^{T-1}\mathbb{E}\left\{\tau(t)\vert\omega\right\}\right]\leq \psi^\text{opt}+\delta,\end{alignat}and\begin{alignat}{1}
		U^\text{min}-\delta\leq\lim_{T\to\infty}\sup \left[\frac{1}{T}\sum_{t=0}^{T-1}\mathbb{E}\left\{U_j(t)\vert\omega\right\}\right]\leq U^\text{max}+\delta.\label{euq3:lemma}\end{alignat}For a integer $T_0$, it can be derived that
	\begin{alignat}{1}
			\frac{1}{T_0}{\sum}_{t=0}^{T_0-1}\mathbb{E}\left\{\tau(t)\vert\omega\right\}\leq \psi^\text{opt}+\delta,\label{euq13:theorem}\end{alignat}and\begin{alignat}{1}
			U^\text{min}-\delta\leq\frac{1}{T_0}{\sum}_{t=0}^{T_0-1}\mathbb{E}\left\{U_j(t)\vert\omega\right\}\leq U^\text{max}-\delta.\label{euq91:theorem}
	\end{alignat}From \cite{DBLP:series/synthesis/2010Neely}, we can note that there exists an i.i.d. policy $\Omega$ such that
	\begin{alignat}{1}\label{euq7:theorem_1}\frac{1}{T_0}\sum_{t=0}^{T_0-1}\mathbb{E}\left\{\left.\tau(t)\right\vert\omega\right\}=\mathbb{E}\left\{\left.\tau(t)\right\vert\Omega\right\},
	\end{alignat}and\begin{alignat}{1}\label{euq7:theorem_2}
	\frac{1}{T_0}\sum_{t=0}^{T_0-1}\mathbb{E}\left\{\left.U_j(t)\right\vert\omega\right\}=\mathbb{E}\left\{\left.U_j(t)\right\vert\Omega\right\}.
\end{alignat}Thus, by plugging (\ref{euq7:theorem_1}) and (\ref{euq7:theorem_2}) into (\ref{euq13:theorem}) and (\ref{euq91:theorem}), we have \textbf{Lemma} \ref{lemma_appendixc}.
\end{IEEEproof}

Next, from \textbf{Lemma} \ref{lemma1}, we have\begin{alignat}{1}\label{euq12:theorem}
	\Delta_V(t)&\leq \sum_{j\in \mathcal{J}}\mathbb{E}\Big\{V\tau(t)+Q_j(t)\left(U^\text{min}-U_j(t)\right)\nonumber\\&+S_j(t)\Big(U_j(t)-U^\text{max}\Big)\Big\vert\boldsymbol{Q}(t),\boldsymbol{S}(t),\Omega\Big\}+H.
\end{alignat}Plugging (\ref{euq1:lemma}) and (\ref{euq3:lemma}) into the right-hand-side of (\ref{euq12:theorem}), letting $\delta\to 0$, and taking expectation of both sides, we have\begin{alignat}{1}\label{equ4:theorem1}
		&\mathbb{E}\left\{L(t+1)-L(t)\vert\boldsymbol{Q}(t)\right\}+V\mathbb{E}\{\tau(t)\vert\boldsymbol{Q}(t)\}\nonumber\\&\leq H+V\psi^\text{opt}.
\end{alignat}By summing up (\ref{equ4:theorem1}) form $t=0$ to $T-1$, and dividing both sides by $T$ and $V$, we have \begin{alignat}{1}\frac{\sum_{t=1}^T\tau(t)}{T}\leq \psi^\text{opt}+\frac{H}{V}+\frac{\mathbb{E}\{L(0)-L(T)\}}{VT},
\end{alignat}which concludes the proof of (\ref{equ2:theorem1}).

Next, from (\ref{equ4:theorem1}), it can be derived that\begin{alignat}{1}\label{equ55}\Delta(t)\leq H+V(\psi^\text{opt}-\tau^\text{min}),
\end{alignat}where\begin{alignat}{1}&\tau^\text{min}=\max_{m\in\mathcal{M}}\left\{\frac{\sum_{n\in\mathcal{N}}a_{n,m}\overline{D}_n\sum_{l=1}^{L_n}\chi_n^{l}}{\max\left\{\max\limits_{m\in\mathcal{M}}\left\{\phi_m^\text{G} f_m^\text{G}\right\},\max\limits_{m\in\mathcal{M}}\left\{\phi_m^\text{A}\right\}\max\limits_{j\in\mathcal{J}}\left\{f_j^\text{max}\right\}\right\}}\right.\nonumber\\&\left.+\frac{\sum_{n\in\mathcal{N}}a_{n,m}\overline{D}_n\min_{l\in\mathcal{L}_n}\{o_n^l\}}{B\log\left(1+\frac{P_m\overline{H_m}}{\overline{\eta_m}+N_0B}\right)}\right\}-\frac{\ln(1-p_0)}{\max_{j\in\mathcal{J}}\{f_j^\text{max}\}}\left(\sum_{j\in\mathcal{J}}e^{\beta}\right.\nonumber\\&\left.\times e^{\alpha g\left(\sum_{m\in\mathcal{M}}\sum_{n\in\mathcal{N}}b_{m,j}a_{n,m}\overline{D}_n\sum_{l=1}^{L_n}\chi_n^l\right)}\right)^{-1}.
\end{alignat}By summing up (\ref{equ55}) from $t=0$ to $T-1$, taking expectations, dividing both sides by $T$, and recalling that $L(t)=\frac{1}{2}\sum_{j\in\mathcal{J}}\left(\left(Q_j(t)\right)^2+\left(S_j(t)\right)^2\right)$, it can be derived that\begin{alignat}{1}&\sum_{j\in \mathcal{J}}\frac{\mathbb{E}\left\{\left(Q_j(T)\right)^2+\left(S_j(T)\right)^2\right\}}{T}\nonumber\\&\leq H + V(\psi^\text{opt}-\tau^\text{min})+\sum_{j\in \mathcal{J}}\frac{\mathbb{E}\left\{\left(Q_j(0)\right)^2+\left(S_j(0)\right)^2\right\}}{T}.
\end{alignat}Thus, for each AP, we have\begin{alignat}{1}\label{equ:108_1} \frac{\mathbb{E}\left\{\left(Q_j(T)\right)^2\right\}}{T}&\leq H + V(\psi^\text{opt}-\tau^\text{min})\nonumber\\&+\sum_{j\in \mathcal{J}}\frac{\mathbb{E}\left\{\left(Q_j(0)\right)^2+\left(S_j(0)\right)^2\right\}}{T},
\end{alignat}and\begin{alignat}{1}\label{equ:108_2}\frac{\mathbb{E}\left\{\left(S_j(T)\right)^2\right\}}{T}&\leq H + V(\psi^\text{opt}-\tau^\text{min})\nonumber\\&+\sum_{j\in \mathcal{J}}\frac{\mathbb{E}\left\{\left(Q_j(0)\right)^2+\left(S_j(0)\right)^2\right\}}{T}.
\end{alignat}By dividing both sides of (\ref{equ:108_1}) and (\ref{equ:108_2}) by $T$, and taking the square root of both sides, we have\begin{alignat}{1}\label{equ19:theorem_1}
		\frac{\mathbb{E}\{Q_j(T)\}}{T}&\leq\left( \frac{H+V\left(\psi^\text{opt}-\tau^\text{min}\right)}{T}\right.\nonumber\\&\left.+\sum_{j\in \mathcal{J}}\frac{\mathbb{E}\left\{\left(Q_j(0)\right)^2+\left(S_j(0)\right)^2\right\}}{T^2}\right)^{\frac{1}{2}},
\end{alignat}and\begin{alignat}{1}\label{equ19:theorem_2}
\frac{\mathbb{E}\{S_j(T)\}}{T}&\leq\left( \frac{H+V\left(\psi^\text{opt}-\tau^\text{min}\right)}{T}\right.\nonumber\\&\left.+\sum_{j\in \mathcal{J}}\frac{\mathbb{E}\left\{\left(Q_j(0)\right)^2+\left(S_j(0)\right)^2\right\}}{T^2}\right)^{\frac{1}{2}}.
\end{alignat}From (\ref{q1}) and (\ref{q2}), it can be derived that\begin{equation}\label{equ17:theorem1}
	Q_j(t+1)\ge Q_j(t)-U_j(t)+U^\text{min},
\end{equation}and\begin{equation}\label{equ17:theorem2}
	S_j(t+1)\ge S_j(t)+U_j(t)-U^\text{max}.
\end{equation}Note that $\mathbb{E}\{Q_j(0)\}<\infty$ and $\mathbb{E}\{S_j(0)\}<\infty$. By summing up (\ref{equ17:theorem1}) and (\ref{equ17:theorem2}) from $t=0$ to $T-1$, taking expectations, and dividing both sides by $T$, it can be derived that\begin{equation}
\frac{\mathbb{E}\{Q_j(T)\}}{T}\ge U^\text{min}-\frac{1}{T}{\sum}_{t=0}^{T-1}U_j(t),
\end{equation}and\begin{equation}
\frac{\mathbb{E}\{S_j(T)\}}{T}\ge \frac{1}{T}{\sum}_{t=0}^{T-1}U_j(t)-U^\text{max}.
\end{equation}Thus, from (\ref{equ19:theorem_1}) and (\ref{equ19:theorem_2}), we have\begin{alignat}{1}
		\frac{1}{T}\sum_{t=0}^{T-1}U_j(t)&\ge U^\text{min}-\left( \frac{H+V\left(\psi^\text{opt}-\tau^\text{min}\right)}{T}\right.\nonumber\\&\left.+\sum_{j\in \mathcal{J}}\frac{\mathbb{E}\left\{\left(Q_j(0)\right)^2+\left(S_j(0)\right)^2\right\}}{T^2}\right)^{\frac{1}{2}},
\end{alignat}and\begin{alignat}{1}
\frac{1}{T}\sum_{t=0}^{T-1}U_j(t)&\leq U^\text{max}+\left( \frac{H+V\left(\psi^\text{opt}-\tau^\text{min}\right)}{T}\right.\nonumber\\&\left.+\sum_{j\in \mathcal{J}}\frac{\mathbb{E}\left\{\left(Q_j(0)\right)^2+\left(S_j(0)\right)^2\right\}}{T^2}\right)^{\frac{1}{2}}.
\end{alignat}This concludes the proof of (\ref{equ3:theorem1}) and (\ref{equ5:theorem1}).
\end{document}